\documentclass[journal]{IEEEtran}
\ifCLASSINFOpdf
\else
\fi
\usepackage[dvipsnames]{xcolor}

\usepackage{times}
\usepackage{epsfig}
\usepackage{graphicx}
\usepackage{amsmath}
\usepackage{amssymb}
\graphicspath{ {./Figures/} }

\begin{document}

%
\title{UVid-Net: Enhanced Semantic Segmentation of UAV Aerial Videos by Embedding Temporal Information}
%
%
%

\author{Girisha S $^\dag$, ~\IEEEmembership{Member,~IEEE,}
         Ujjwal Verma $^\dag$, ~\IEEEmembership{Senior Member,~IEEE,}
        Manohara Pai M M *,~\IEEEmembership{Senior Member,~IEEE,}
        ~and Radhika M Pai ,~\IEEEmembership{Senior Member,~IEEE,} 
\thanks{* Corresponding Author}
\thanks{\dag ~ Equal Contribution}
\thanks{Girisha S, Manohara Pai and Radhika Pai are with Department of Information and Communication Technology, Manipal Institute of Technology, Manipal Academy of Higher Education, Manipal, India. Ujjwal Verma is with Department of Electronics and Communication Engg, Manipal Institute of Technology, Manipal Academy of Higher Education, Manipal, India.}}

%
%

\markboth{Journal of \LaTeX\ Class Files,~Vol.~13, No.~9, September~2014}%
{Shell \MakeLowercase{\textit{et al.}}: Bare Demo of IEEEtran.cls for Journals}
%



\maketitle

\begin{abstract}
Semantic segmentation of aerial videos has been extensively used for decision making in monitoring environmental changes, urban planning, and disaster management. The reliability of these decision support systems is dependent on the accuracy of the video semantic segmentation algorithms. The existing CNN based  video semantic segmentation methods have enhanced the image semantic segmentation methods by incorporating an additional module such as LSTM or optical flow for computing temporal dynamics of the video  which is a computational overhead. The proposed research work modifies the CNN architecture by incorporating temporal information to improve the efficiency of video semantic segmentation. 

In this work, an enhanced encoder-decoder based CNN architecture (UVid-Net) is proposed for UAV video semantic segmentation. The encoder of the proposed architecture embeds temporal information for temporally consistent labelling. The decoder is enhanced by introducing the feature-refiner module, which aids in accurate localization of the class labels. The proposed UVid-Net architecture for UAV video semantic segmentation is quantitatively evaluated on extended ManipalUAVid  dataset. The performance metric mIoU of 0.79 has been observed which is significantly greater than the other state-of-the-art algorithms. Further, the proposed work produced promising results even for the pre-trained model of UVid-Net on urban street scene with fine tuning the final layer on UAV aerial videos.




\end{abstract}

\begin{IEEEkeywords}
UAV video, semantic segmentation, transfer learning, U-Net, Deep Learning
\end{IEEEkeywords}

%
\IEEEpeerreviewmaketitle

\section{Introduction}
%
%
%
%
\IEEEPARstart{T}{he} analysis of data collected from airborne sensors such as aerial images/videos are increasingly becoming a vital factor in many applications such as scene understanding, studying the ecological variations  \cite{pai2019automatic}, \cite{River-Ice-Segmentation2020}, \cite{pai2020improved} tracking of vehicles/animals/humans \cite{38},  \cite{40}, \cite{35},  surveying the urban development \cite{42}, \cite{BuildingSegmentation2020}, \cite{RoadSegmentation2020} etc. Besides, aerial image analysis has been used for assessing the damage immediately after a natural disaster \cite{43}. Typically, the aerial images are captured by different imaging modalities such as Synthetic Aperture Radar (SAR) \cite{37}, hyper-spectral imaging \cite{39} which are present on-board a satellite. Recently, the Unmanned Aerial Vehicles (UAV) have also been widely used for various applications such as disaster management, urban planning, tracking of wildlife, agricultural planning etc \cite{bergsma2019operational}, \cite{bulatov2011context}. Due to rapid deployment and a customized flight path, the UAV images/videos, could provide additional finer details and complement satellite-based image analysis approaches for critical applications such as disaster response \cite{luo2019unmanned}. Besides, the UAV images could be utilized along with satellite images for better urban planning or geographical information updating. Typically, the UAV image/video analysis is limited for object detection \cite{fang2011improved}, \cite{7938673} and recognition \cite{wang2012framework} tasks such as building detection, road segmentation etc. However, to the best of our knowledge, there are limited works on semantic segmentation of UAV images or videos \cite{semantic}, \cite{wang2019deep}.
\begin{figure}[tbp]
	\begin{center}
		\includegraphics[width=\columnwidth,height=2.5in]{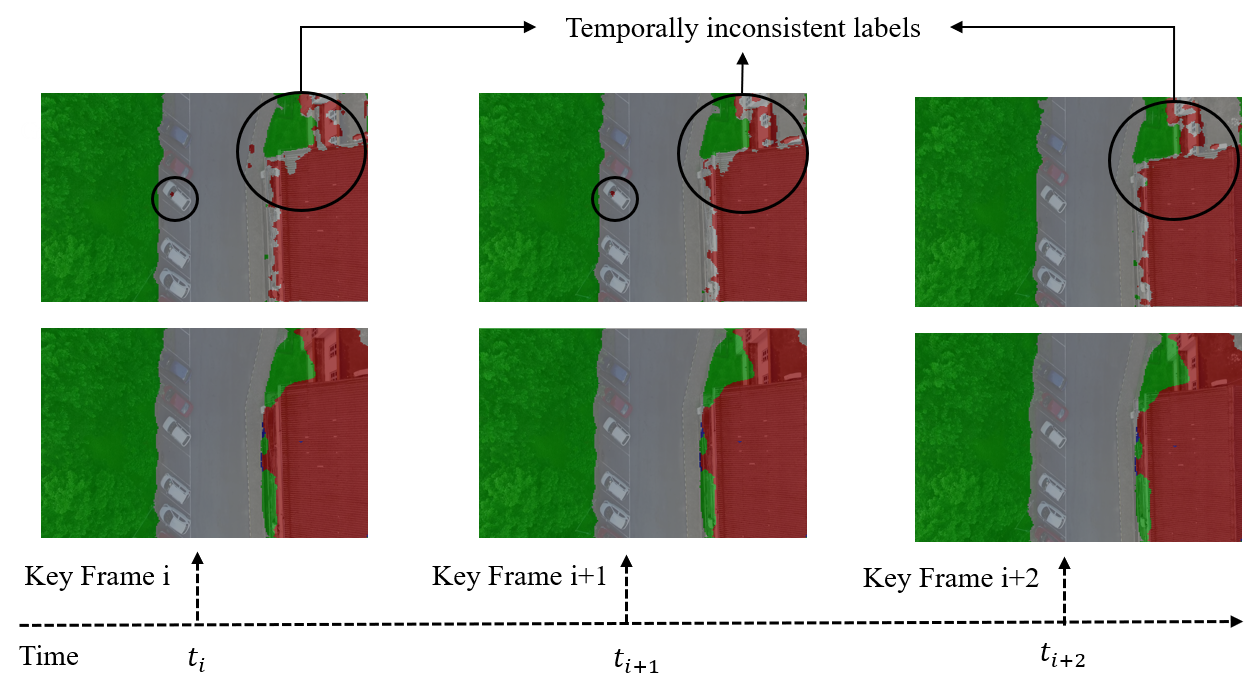}
	\end{center}
	
	\caption{ Importance of temporal consistency in video scene labeling: First row represents the temporal inconsistent labels produced by image segmentation algorithm (U-Net) on videos. Second row depicts an example of temporally consistent labelling obtained using the proposed approach (UVid-Net).}
	\label{fig:1}
	
\end{figure}
\par Segmentation is a crucial task for scene understanding and has been used for various applications \cite{18}, \cite{2}, \cite{Ujjwal-Seg1}. Semantic segmentation is a process of assigning predetermined class labels to all the pixels in an image. Semantic segmentation of an \textit{image} is a widely studied topic in computer vision. However, the extension of semantic segmentation for video applications is a non-trivial task. One of the challenges in video semantic segmentation is to find a way to incorporate temporal information. Figure \ref{fig:1} illustrates the importance of temporal information in the context of video acquired by UAV. The poor segmentation in the greenery class can be observed in the $(i+1)^{th}$ keyframe which can be improved by embedding temporal information from the past frames. 

\par In a typical video semantic segmentation approach, a sequential model is added on top of the frame-wise semantic segmentation module, thus creating an overhead \cite{13}. Besides, features/label propagation \cite{7}, which re-utilizes features/labels from previous frames has also been utilized to capture the temporal information. However, these methods depend on the establishment of pixel correspondence between two frames. Recently, video prediction based approach \cite{22} has been used to generate new training images and has achieved state-of-the-art performance for video semantic segmentation. However, this approach uses an additional video prediction model to learn the motion information. 

\par This work focuses on semantic segmentation of videos acquired using UAV. The proposed method demonstrates that a simple modification in the encoder branch of CNN is able to capture the temporal information from the video thus eliminating the need for an extra sequential model for computing correspondence for feature/label propagation. 

\par A new encoder-decoder based CNN architecture (UVid-Net) proposed in this work has two parallel branches of CNN layers for feature extraction. This new encoding path captures the temporal dynamics of the video by extracting features from multiple frames.  These features are further processed by the decoder for the estimation of class labels. The proposed algorithm utilizes a new decoding path that retains the features of encoder layers for decoders. The contribution of the paper can be summarized as,
\begin{itemize}
    \item The dependence of existing methods on optical flow/ConvLSTM for the establishment of temporal correspondence is an overhead for video semantic segmentation. Hence, a new encoding path is presented consisting of two parallel branches for extracting temporal and spatial features for video semantic segmentation. This new encoding path eliminates the need for an extra sequential module (ConvLSTM) or computation of optical flow for establishing temporal correspondence. 
    
    \item  A modified up-sampling path is proposed which uses a feature-refiner module to capture fine-grain features for accurate classification of class boundary pixels. The feature-refiner module also reduces the number of parameters (11.68\% reduction) and the computational complexity (11\% reduction) of the model as compared to the traditional decoder module.

	\item An extended version of UAV video semantic segmentation dataset is presented. This dataset is an extension of ManipalUAVid dataset \cite{19} and contains additional videos captured at new locations. Fine pixel-level annotations are provided for four background classes namely greenery, roads, constructions and water bodies as per the policy adopted in \cite{19}. The dataset is available for download at https://github.com/uverma/ManipalUAVid 
	\item This work also studies the performance of the proposed UVid-Net trained on a larger urban street scene dataset for semantic segmentation for segmentation of UAV aerial videos. The capability of UVid-Net to utilize transferable features allows the model to be retrained with a few labelled data.
	
\end{itemize}

\par This paper is organized as follows: Section \ref{sec:literature} summarizes the recent developments in video semantic segmentation. Section \ref{sec:methodology} describes the architecture of the proposed network UVid-Net and Section \ref{Sec: results} presents the various results obtained. 

\section{Related works}
\label{sec:literature}
Video semantic segmentation is generally addressed by utilizing traditional energy-based algorithms such as CRF or deep learning-based algorithms such as CNN, RNN, LSTM, etc. One of the challenges in video semantic segmentation is to embed temporal information. Learning the dynamics of the video aids in improving the performance of video semantic segmentation by ensuring temporal consistency. Despite this interest, previous works such as \cite{19}, \cite{semantic}, \cite{1} extended the traditional image semantic segmentation approach for video semantic segmentation. These approaches segment all the frames independently of each other which fails to capture the dynamics of the video. Recent advances in video semantic segmentation by utilizing Spatio-temporal information can be categorized into roughly two groups: Deep Learning based methods and CRF based methods.
\begin{figure*}[h]
	\begin{center}
		\includegraphics[width =0.8\textwidth]{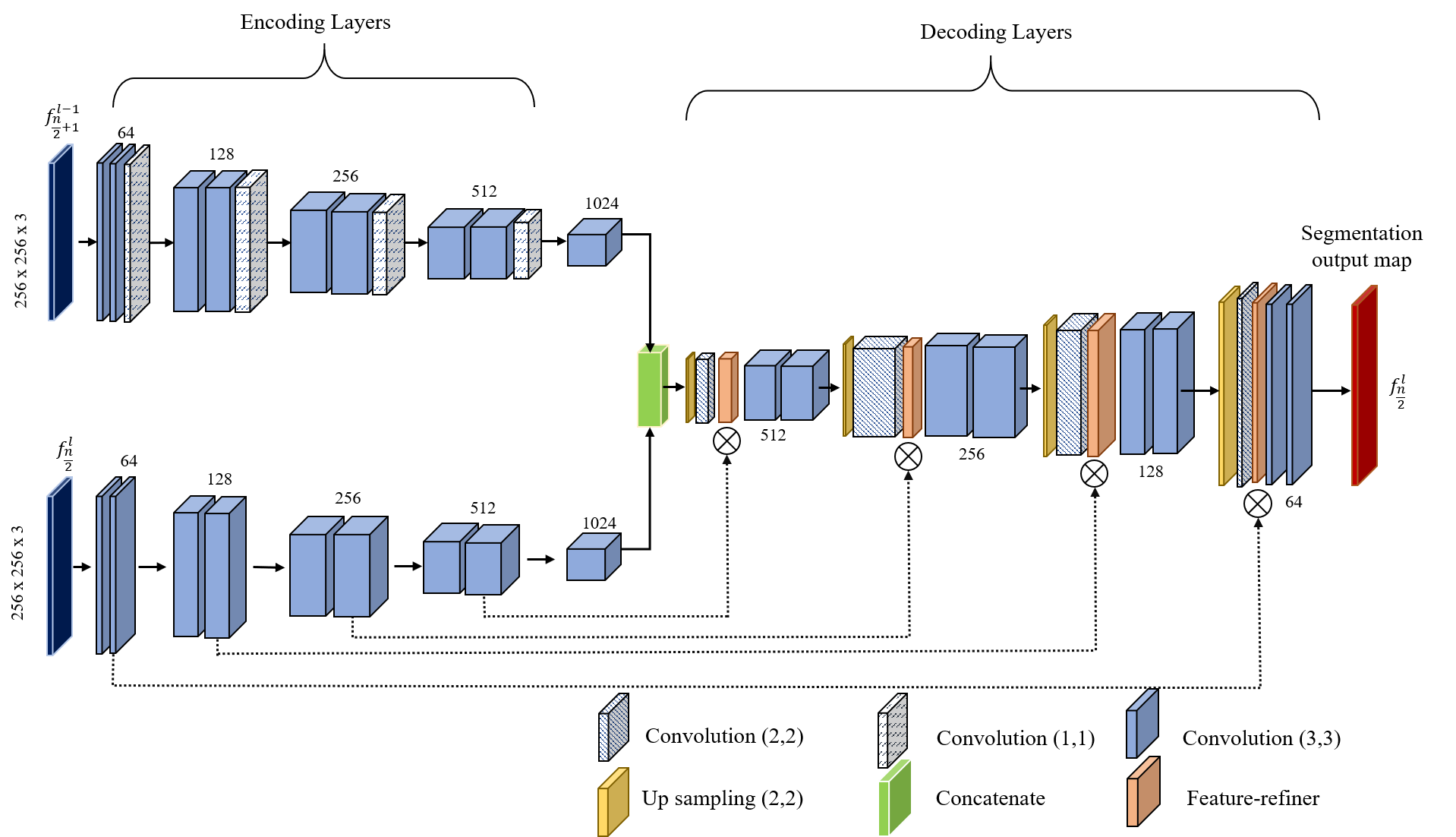}
	\end{center}
	\caption{UVid-Net: Overview of the proposed architecture for UAV video semantic segmentation (U-Net encoder). The architecture consists of encoding path to extract spatio-temporal features and an upsampling path which produces smoother segmentation boundaries.}
	\label{fig:unet}
\end{figure*}

There exists several CNN based semantic segmentation approaches in literature such as \cite{26}, \cite{24}, \cite{22}, \cite{23},  \cite{29} etc. The authors of \cite{26}, proposed bilateral segmentation network (BiSeNet) to capture spatial and contextual information for semantic segmentation. In a separate study \cite{liu2021part}, the authors used part-object relationship for a robust salient object segmentation.  In \cite{lian2021cascaded}, authors used multiple ASPP module to increase the density of sampling distribution. However, these approaches analyse single image, while the proposed work aims to incorporate temporal information for video semantic segmentation.


Popular CNN based algorithms like \cite{2}, \cite{4} used encoder and decoder based architecture for learning the various patterns of the data and localizing the class labels. These algorithms are dependent on a large densely annotated dataset. However, obtaining a finely annotated large dataset is expensive, time-consuming and challenging. To address the issue of limited training data, GANs were utilized \cite{24}. Few authors (\cite{28}, \cite{15}) used GAN to learn the dynamics of the video and perform video scene parsing. GAN can be trained to parse future frames as well as label images as proposed by \cite{15}.

Besides, temporal dynamics are also learnt using a sequential model like LSTM \cite{wang2019deep}. Moreover,  LSTM is also used to select keyframes for video scene parsing \cite{16}. Bowen Wang et al. \cite{wang2020noisy}, proposed Noisy-LSTM which uses convLSTM for video semantic segmentation. The strategy used is to train network with noisy images to disrupt the temporal information. Recently, memory modules are also explored for learning the temporal dynamics of the video \cite{paul2021local}. Few authors explored the attention mechanism with CNN to perform video semantic segmentation \cite{27}, \cite{wang2021temporal}. Hao Wang et al. \cite{wang2021temporal} proposed TMANet which uses attention mechanism to capture long range temporal information required for video semantic segmentation. In another study \cite{xuerui2020lsmvos}, authors used pixel-level matching between two consecutive frames to obtain global and local similarity maps for video object segmentation. However, it is challenging to determine the attention coefficients.  Optical flow is another popular choice for the establishment of temporal correspondence between two consecutive frames \cite{25}. Few studies such as \cite{22} and \cite{23} proposed to predict labels and images jointly to efficiently train deep learning models with less training data. However, the dependence of deep learning algorithms on large annotated datasets limits the development of deep learning algorithms for other contexts such as UAV, etc. 


Many researchers have explored Conditional Random Field (CRF) for incorporating Spatio-temporal information in video semantic segmentation. CRF is a graphical model that captures a large spatial relationship between pixels. Hence it is widely used in literature for context-aware scene parsing \cite{3}, \cite{6}. 
 CRF can be extended to incorporate temporal information as shown in few literatures such as \cite{5}, \cite{6}, \cite{8}, \cite{9}, \cite{10}, \cite{12}, but it depends on the reliable estimation of temporal links. The authors of \cite{5} utilized 3D CRF along with optimized feature space for video semantic segmentation. However, 3D CRF is impractical for videos since its computationally expensive. Potential energies based on temporal information was also explored for producing temporally consistent labels \cite{9}.Few researchers incorporated CNN with in CRF frame work to obtain initial estimation of class labels \cite{8}, \cite{12}. Authors of \cite{10}, used conditional restricted Boltzmann machine along with CRF to learn the temporal and shape features required for video semantic segmentation. 
In general, optical flow is widely used to establish the temporal link and propagate features and labels. However, estimation of accurate optical flow is an overhead for real-time video semantic segmentation. In the recent work of \cite{girisha2020semantic}, a new potential term was proposed to enhance the temporal smoothness of video semantic segmentation without the usage of optical flow. In an another study, higher-order potential energies were explored for video semantic segmentation \cite{6}. Class labels in CRF are inferred by using an inference algorithm which is computationally intensive and impractical for video processing.

\par The existing state-of-the-art method for video semantic segmentation predicts frames and its labels from the historic data \cite{22}. However, this approach is dependent on a reliable estimation of temporal correspondence between two consecutive frames. Temporal links are generally established by utilizing dense optical flow-based methods \cite{5}. Optical flow estimation is an overhead and accuracy of semantic segmentation depends on the accuracy of optical flow estimation. Besides, the error in optical flow estimation can lead to misaligned predicted labels in the future frames, thus affecting the accuracy of the segmentation. The proposed work eliminates the need for computing optical flow, thus reducing the overhead. 



In this work, a two-branch encoder is proposed for incorporating temporal smoothness in video semantic segmentation. Multi-branch CNNs are popularly used in video processing due to their ability to capture the relationship between the sequence of frames. Several authors used multi-branch CNNs to perform video classification \cite{wang2018appearance}, action recognition \cite{peng2016multi} and video captioning \cite{yu2016video}. Few authors utilized multi-branch CNN architecture to provide attention mechanism. Authors of \cite{qiao2016deep} explored multi-branch CNN to extract features from different frames. In \cite{hu2020temporally}, the authors proposed to utilize multiple shallow networks to extract features from consecutive frames to perform video semantic segmentation. To the best of our knowledge, lightweight multi-branch CNNs are not explored to perform video semantic segmentation of UAV videos.  

\section{Methodology}
\label{sec:methodology}

\par
\begin{figure}[t]
	\begin{center}
		\includegraphics[width=\columnwidth,height=2.5in]{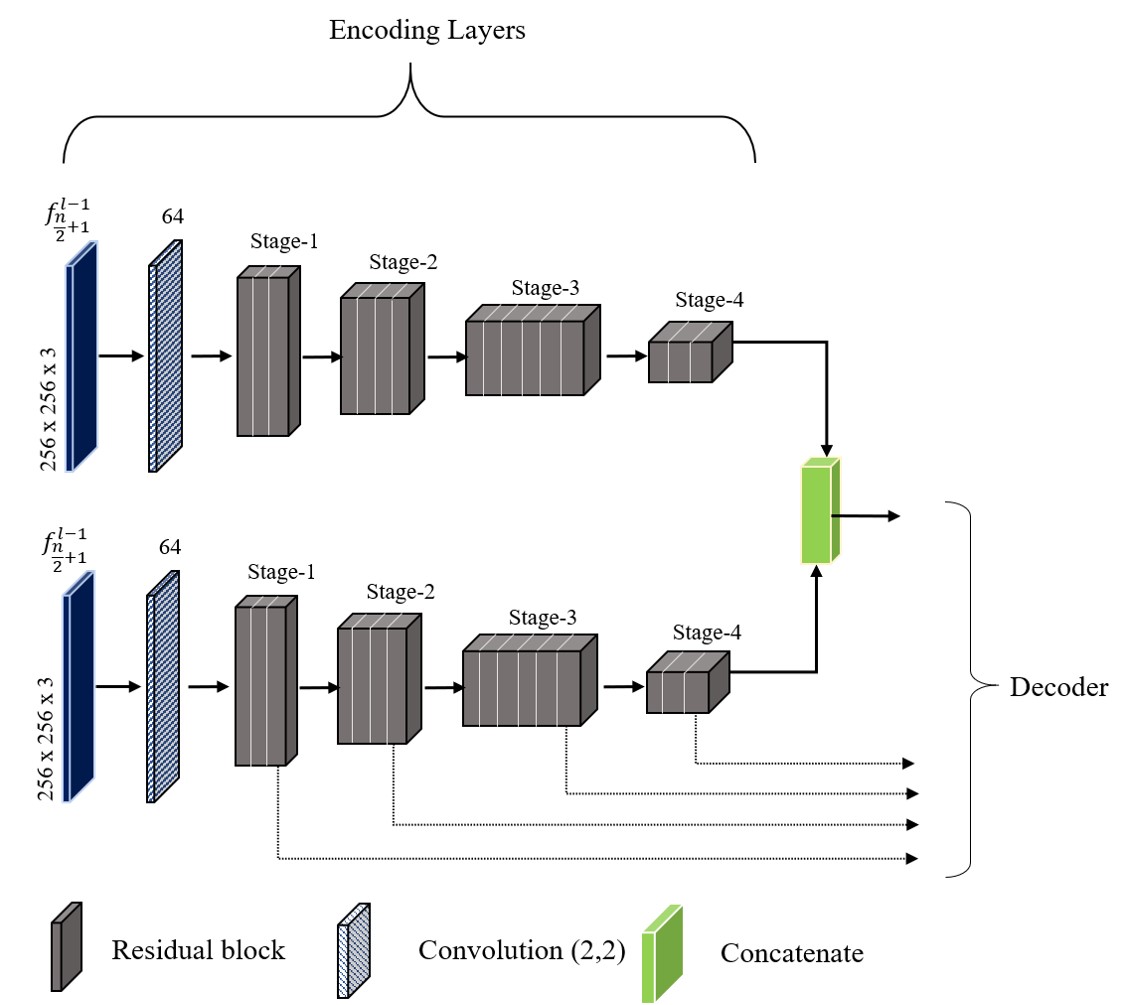}
	\end{center}
	\caption{UVid-Net (ResNet-50 encoder): Overview of the proposed architecture for UAV video semantic segmentation with ResNet-50 encoder.}
	\label{fig:res}
\end{figure}

 This section describes the encoder (Section \ref{SubSec:Encoder}) and decoder module (Section \ref{SubSec:Decoder}) of the proposed approach.  The Figure \ref{fig:unet} and  Figure \ref{fig:res} shows the proposed architecture with U-Net and ResNet-50 feature extractor respectively. In a typical video, the changes between two consecutive frames are very minimum and hence processing every frame is redundant and time-consuming for video semantic segmentation. However,  selecting keyframes at constant interval may result in loss of useful information required for temporal consistency. This would be detrimental for video semantic segmentation methods which depend on temporal features. Hence, in the present study, the keyframes are identified using the shot boundary detection approach presented in \cite{19} (on an average, a shot consists of 15-20 frames). The use of shot boundary detection method for dynamically identifying the keyframes ensures that the frames containing useful information are not ignored.

\par Let us represent the $i^{th}$ frame from the $l^{th}$ shot in a video as $f_i^l$. The inputs to the two branches of UVid-Net (Figures \ref{fig:unet}, \ref{fig:res}) are the two frames from two consecutive shots: $f_{(n/2+1)}^{(l-1)}$ (upper branch) and $f_{n/2}^l$ (lower branch) , where $n$ represents the total number of frames in a shot. These two frames correspond to the next frame after the middle frame of the previous shot  $f_{(n/2+1)}^{(l-1)}$ and the middle frame from the current shot. These two input frames produce the semantic segmentation for the middle frame of the current shot   $f_{n/2}^l$. For the first shot, since there is no prior shot, the first frame ($f_1^1$) of the video and middle frame ($f_{n/2}^1$) of the first shot is considered as input to  the network.
In the rest of this document, the middle frame of a shot is considered as the keyframe, as per the policy followed for UAV video semantic segmentation \cite{19}.

\subsection{Encoder}  
\label{SubSec:Encoder}
\par  
In this work, the performance of two different architectures (U-Net and ResNet-50 encoders) is studied for feature extraction. U-Net encoder consists of a convolutional layer and maxpool layers for feature extraction. The ResNet-50 feature extractor consists of residual blocks which helps in alleviating the vanishing gradient. These two feature extractors are different and comparing their performance on multi-branch CNN helps us in providing insight into the robustness of the model. In the following text, UVid-Net (U-Net encoder) and UVid-Net (ResNet-50 encoder) refer to the proposed architecture with U-Net encoder and ResNet-50 encoder module respectively. It may be noted that the decoder module is identical for both the architectures.

\subsubsection{U-Net Encoder} 
\par The upper branch of the encoder (Figure \ref{fig:unet}) contains four blocks. Each block consists of two consecutive $3\times 3$ convolution layers with batch normalization and ReLU activation function as in the encoder of U-Net \cite{26}. Finally, the activation is then passed through a $1\times 1$ convolution layer which is additionally introduced to reduce the dimensions of the feature maps. Lastly, a maxpooling layer with stride $(2,2)$ is applied to extract the most prominent features for the subsequent layers. As in the traditional U-Net, the number of feature maps doubles after each max pooling operation, starting with 64 feature maps for the first block.  

The lower branch of the encoder also consists of four blocks. Each block in the lower branch has a $3 \times 3$ convolution layer with batch normalization and ReLU activation function and the second set of $3 \times 3$ convolution layer with batch normalization and ReLU activation function. This is followed by a maxpooling layer which extracts most prominent features.  Similar to the upper branch, the number of feature maps doubles after each max pooling operation.

The features extracted by the upper and lower branch of the encoder are fed to two separate bottleneck layers consisting of $3 \times 3 $ convolution with 1024 features maps. Finally, the activation of both these branches is concatenated and fed to the decoder.

\subsubsection{ResNet-50 encoder}
\label{resnet} 
Besides the UNet based encoder described above, the ResNet-50 architecture (Figure \ref{fig:res}) could also be used as a branch in the encoder.  ResNet-50 is a CNN architecture proposed for image classification. This architecture proposed the idea of skipping a few layers to learn identity mapping.  ResNet-50 has also been widely used as a feature extractor for transfer learning applications \cite{he2016deep}.

In the present study, the upper branch and lower branch consists of identical ResNet50 architecture to extract features (Figure \ref{fig:res}). This architecture consists of an initial convolution operation with kernel size (7x7)  followed by batch normalization layer and ReLU activation function. Subsequently, a max pool operation with kernel size (3x3) is applied. Followed by the maxpool operation the architecture consists of four stages. The first stage consists of three residual blocks each containing three layers. Each of these residual blocks consists of 64, 64 and 128 filters. The second stage consists of 4 residual blocks with three layers each. These three layers use 128, 128 and 256 filters. The third stage consists of 6 residual blocks with three layers each. These layers use 256, 256 and 512 filters. The fourth stage consists of 3 residual blocks with three layers each. These layers use 512, 512 and 1024 filters. The first residual blocks of stage 2,3 and 4 utilizes stride operation to reduce the input dimension by 2 in terms of width and height. First and last layers in every residual block consist of (1x1) kernel size and the second layer consists of (3x3) kernel size. All residual block consists of identity connection which solves the vanishing gradient problem. 

The activations of upper and lower ResNet50 branch are concatenated and is further used by the decoder to perform semantic segmentation.   



\subsection{Decoder} 
\label{SubSec:Decoder}
\par  
In an encoder-decoder based architecture, the consecutive max pooling operations in encoder reduces feature maps size and results in the loss of spatial resolution. Hence to compensate for this loss of information, skip connections are popularly used from encoding layers to decoding layers \cite{2},\cite{4}. Networks like U-Net use concatenation operation where the feature maps from the last layer of each block in the encoder are stacked with the feature maps of corresponding decoding layers. Here, we argue that element-wise multiplication of the feature maps from the last layer of each block in encoder with the corresponding decoding layers results in better representation of feature maps. This module which performs element-wise multiplication of feature maps is called as feature-refiner since it \textbf{refines} the features of the corresponding encoding path. In addition to the improvement in segmentation, the proposed feature-refiner module reduces the number of learnable parameters as compared to the concatenation operation.  For instance, the total number of parameters for UVid-Net (U-Net encoder) with multiplication is $23,745,032$ whereas the total number of parameters for UVid-Net (U-Net encoder) with concatenation is $26,878,472$. The experimental results (Section \ref{Sec:Decoder}) show that the element-wise multiplication of the encoder feature map with the corresponding decoder feature map produces a more smoother segmentation map.

As discussed earlier, the decoder module is identical for both UVid-Net (U-Net encoder) and UVid-Net (ResNet-50 encoder) (Figure \ref{fig:unet} and \ref{fig:res}). The decoder path of the proposed architecture contains four blocks. Each of these blocks consists of an upsampling layer with stride 2. This is followed by a convolution layer with filter size ($2,2$). The output of this is passed through a feature-refiner module which multiplies the corresponding feature maps of the encoder (lower branch) and the decoder. Note that the last layer of each stage/block of the lower branch encoder is merged with corresponding decoder layers.  This is followed by convolution layers and the ReLU activation layer. At the last layer, the SoftMax layer is applied to obtain the probability of pixels belonging to each class.

\begin{figure*}[!t]
	
	\begin{tabular}{ccccc}
		\begin{minipage}{40pt}
			\includegraphics[width=1.2in, height=0.74in]{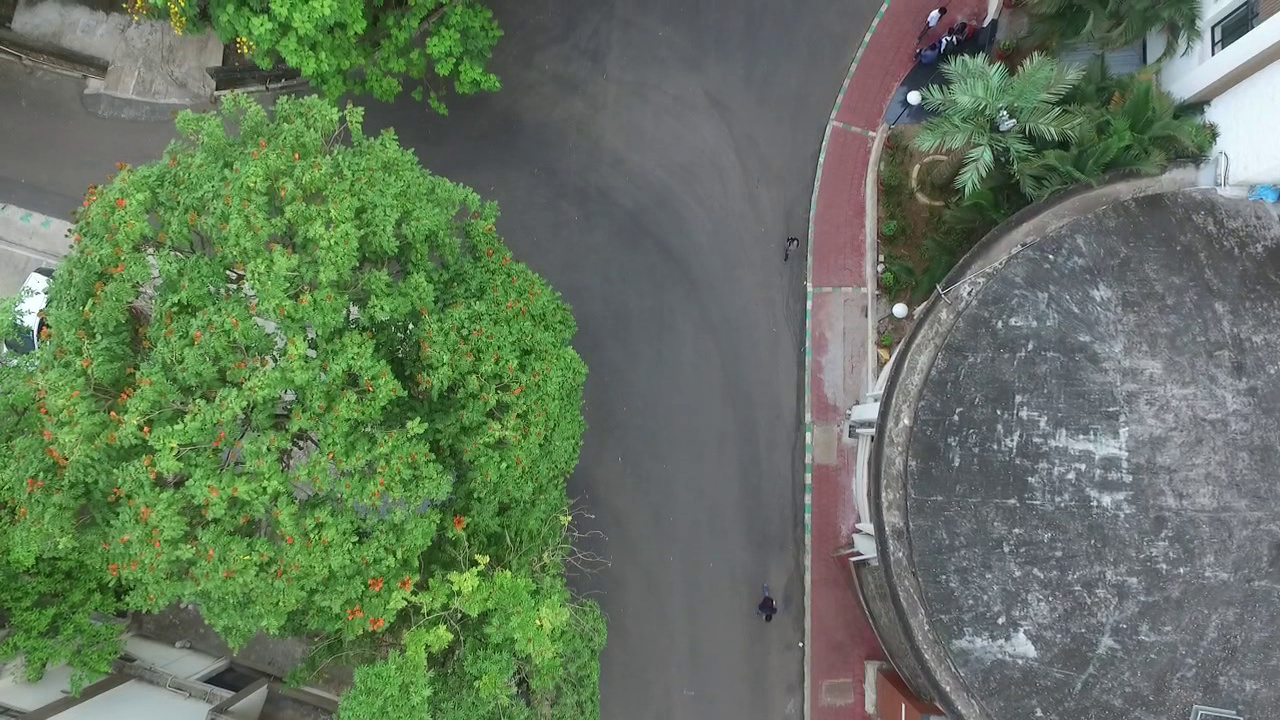}
		\end{minipage}
		&
		\hspace{1.4cm}
		\begin{minipage}{40pt}
			\includegraphics[width=1.2in, height=0.74in]{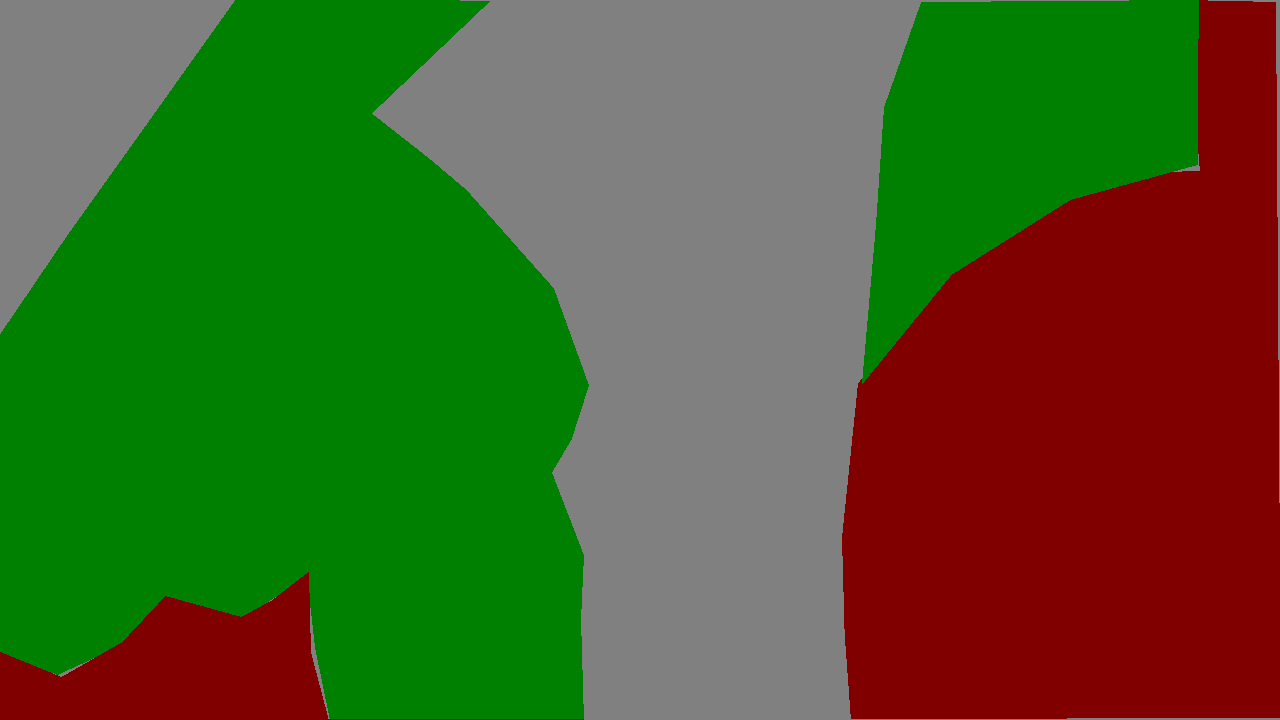}
		\end{minipage}
		&
		\hspace{1.4cm}
		\begin{minipage}{40pt}
			\includegraphics[width=1.2in, height=0.74in]{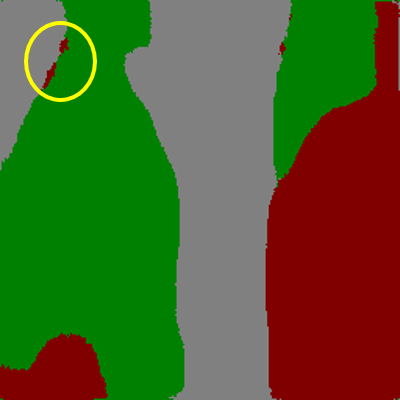}
		\end{minipage}
		&
		
		\hspace{1.4cm}
		\begin{minipage}{40pt}
			\includegraphics[width=1.2in, height=0.74in]{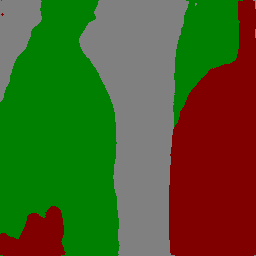}
		\end{minipage}
		&
		\hspace{1.4cm}
		\begin{minipage}{40pt}
			\includegraphics[width=1.2in, height=0.74in]{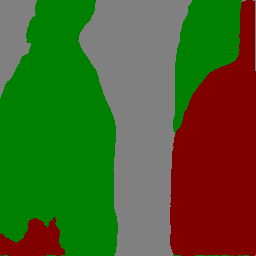}
		\end{minipage}
		\\
		\\
		\begin{minipage}{40pt}
			\includegraphics[width=1.2in, height=0.74in]{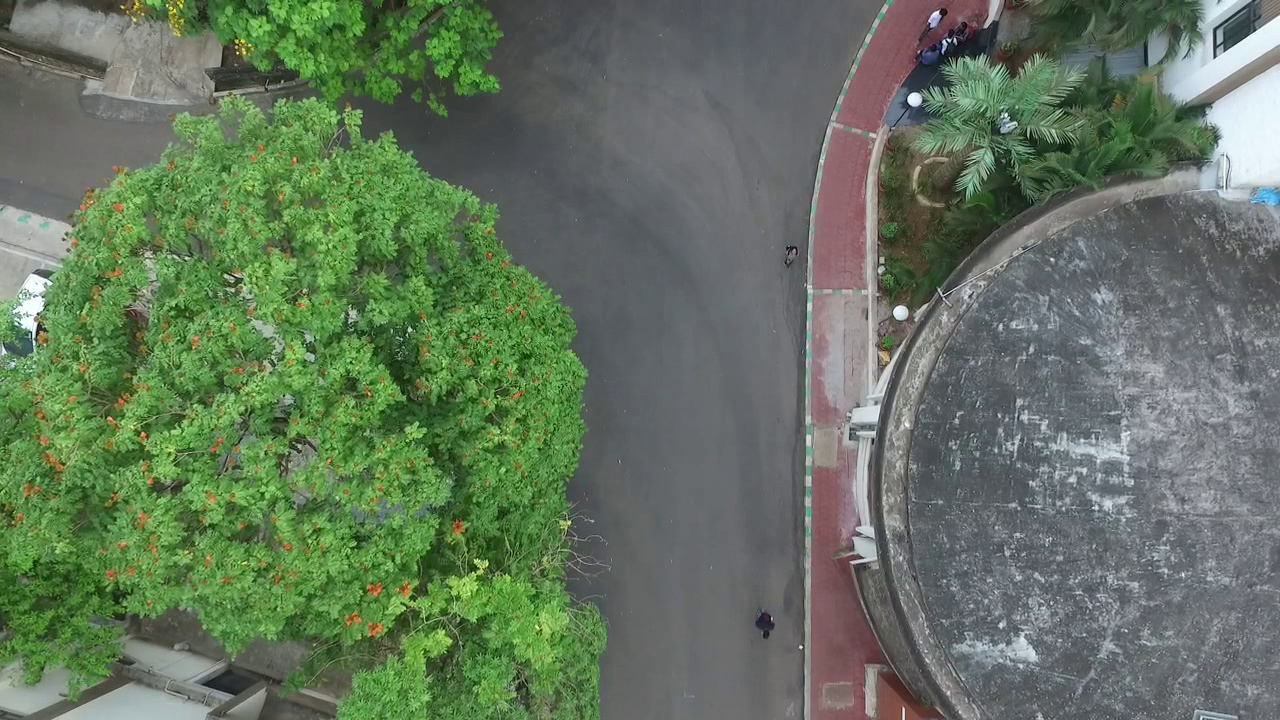}
		\end{minipage}
		&
		\hspace{1.4cm}
		\begin{minipage}{40pt}
			\includegraphics[width=1.2in, height=0.74in]{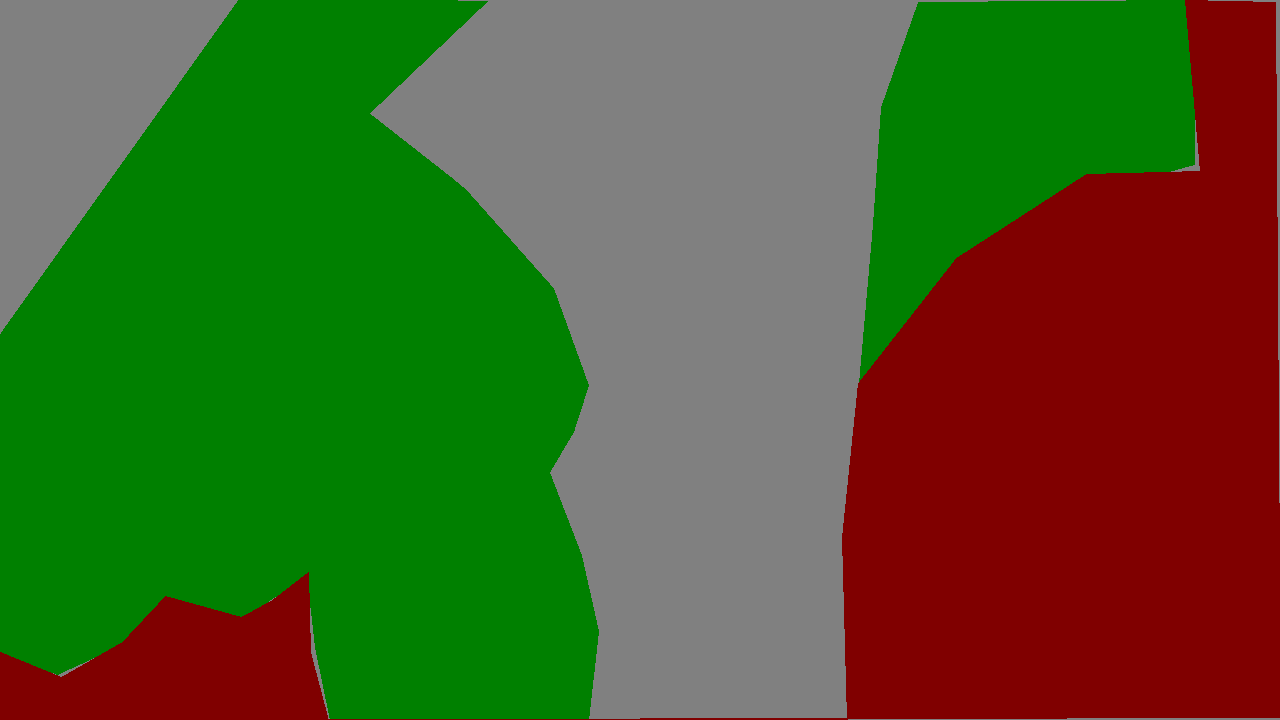}
		\end{minipage}
		&
		\hspace{1.4cm}
		\begin{minipage}{40pt}
			\includegraphics[width=1.2in, height=0.74in]{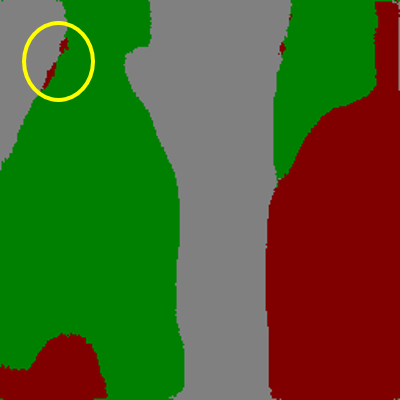}
		\end{minipage}
		
		&
		\hspace{1.4cm}
		\begin{minipage}{40pt}
			\includegraphics[width=1.2in, height=0.74in]{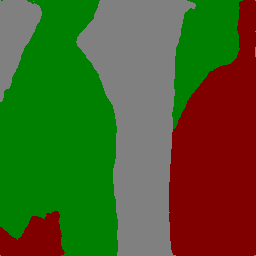}
		\end{minipage}
		&
		\hspace{1.4cm}
		\begin{minipage}{40pt}
			\includegraphics[width=1.2in, height=0.74in]{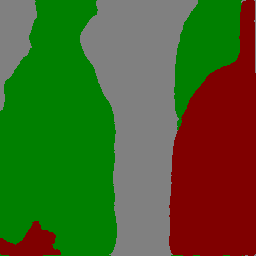}
		\end{minipage}
		\\
		\\
		\begin{minipage}{40pt}
			\includegraphics[width=1.2in, height=0.74in]{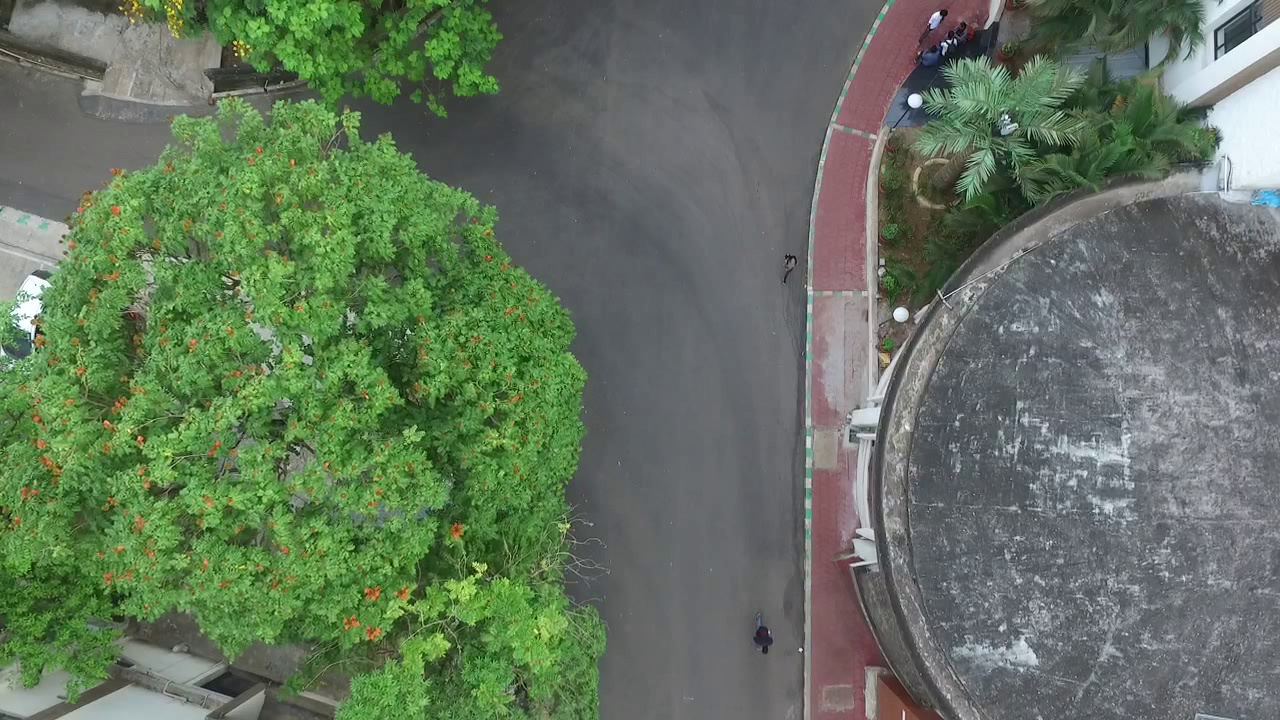}
		\end{minipage}
		&
		\hspace{1.4cm}
		\begin{minipage}{40pt}
			\includegraphics[width=1.2in, height=0.74in]{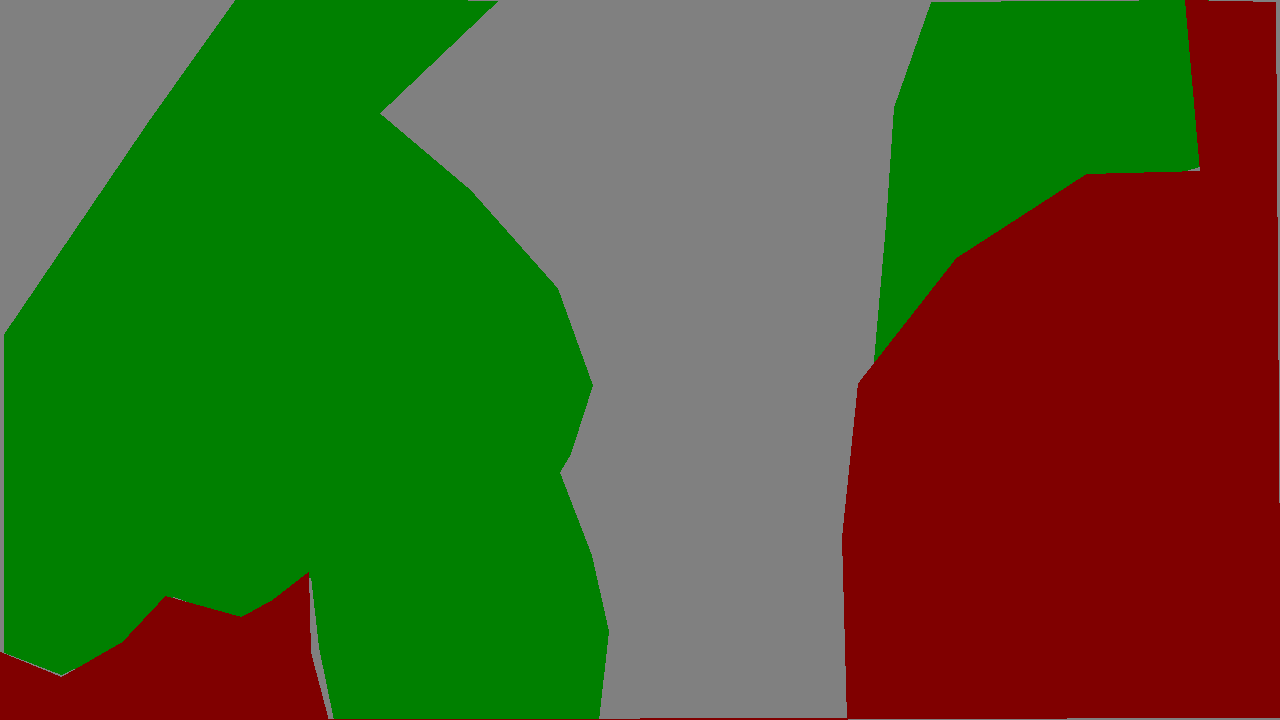}
		\end{minipage}
		&
		\hspace{1.4cm}
		\begin{minipage}{40pt}
			\includegraphics[width=1.2in, height=0.74in]{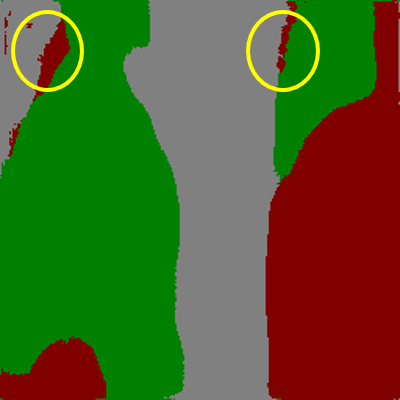}
		\end{minipage}
		
		&
		\hspace{1.4cm}
		\begin{minipage}{40pt}
			\includegraphics[width=1.2in, height=0.74in]{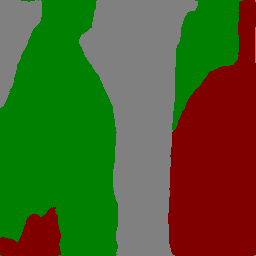}
		\end{minipage}
		&
		\hspace{1.4cm}
		\begin{minipage}{40pt}
			\includegraphics[width=1.2in, height=0.74in]{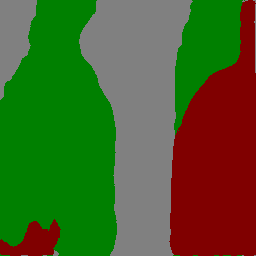}
		\end{minipage}
		\\
		\\
		\begin{minipage}{40pt}
			\includegraphics[width=1.2in, height=0.74in]{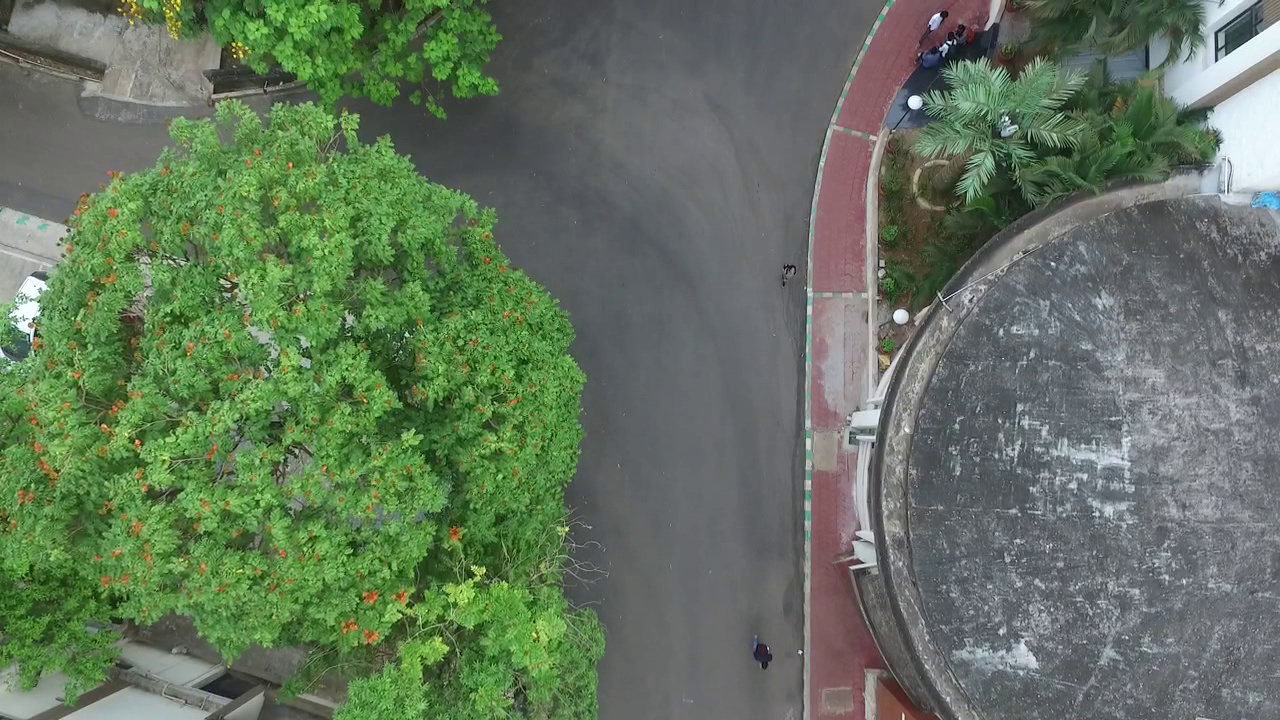}
			\centering{(a)}
		\end{minipage}
		&
		\hspace{1.4cm}
		\begin{minipage}{40pt}
			\includegraphics[width=1.2in, height=0.74in]{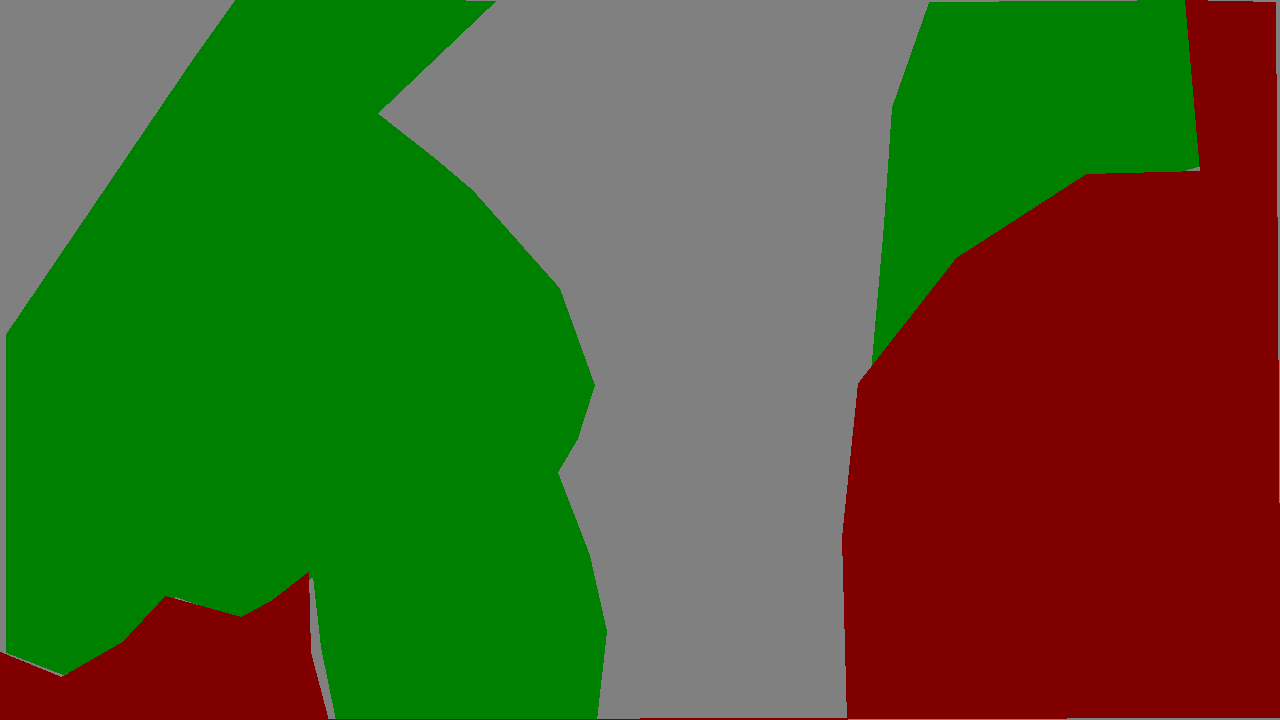}
			\centering{(b)}
		\end{minipage}
		&
		\hspace{1.4cm}
		\begin{minipage}{40pt}
			\includegraphics[width=1.2in, height=0.74in]{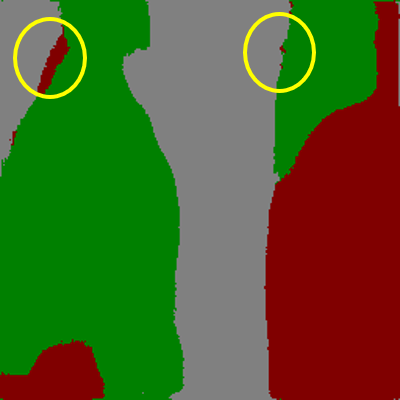}
			\centering{(c)}
		\end{minipage}
		
		&
		\hspace{1.4cm}
		\begin{minipage}{40pt}
			\includegraphics[width=1.2in, height=0.74in]{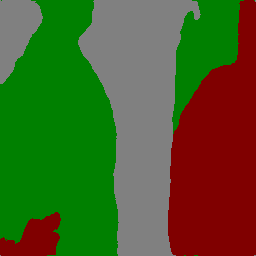}
			\centering{(d)}
		\end{minipage}
		&
		\hspace{1.4cm}
		\begin{minipage}{40pt}
			\includegraphics[width=1.2in, height=0.74in]{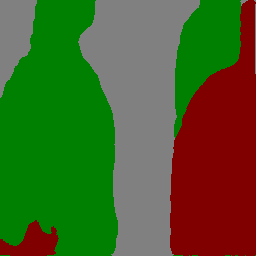}
			\centering{(e)}
		\end{minipage}
		\\
		\\
		
	\end{tabular}
	\caption{Comparing the performance of the proposed two branch encoder module with single branch encoder: Column (a) shows four consecutive keyframes and column (b) shows its corresponding ground-truth images. Column (c) shows the results of single branch encoder viz. U-Net, while column (d) and (e) shows the results of two branch encoder architectures viz. UVid-Net (with U-Net encoder) and UVid-Net (ResNet50 encoder) respectively. Yellow circles highlights the temporal inconsistency produced by single branch U-Net in semantic segmentation.  Here, green, gray, red and blue  colour represents the greenery, road, construction and water bodies class respectively. }
	\label{fig:temporal}
	
\end{figure*}

\section{Results and discussion}
\label{Sec: results}


\par In the present study, an extended version of ManipalUAVid \cite{19} dataset is used to evaluate the performance of the UVid-Net for UAV video semantic segmentation. The proposed architecture is trained by utilizing categorical cross-entropy loss with Adam optimizer for learning the parameters of the model.  In this section, it is shown experimentally that the proposed encoder module is able to incorporate temporal smoothness for video semantic segmentation (Section \ref{Sec:TempInfo}). Further, the effectiveness of the feature-refiner in the decoder module is demonstrated in Section \ref{Sec:Decoder}. Finally, the performance of the proposed architecture is compared with the state-of-the-art methods for video semantic segmentation (Section \ref{Sec:SOTAComp}).

\subsection{Dataset: ManipalUAVid}
\par This paper presents an extended version of ManipalUAVid \cite{19} dataset for semantic segmentation of UAV videos. This extended dataset consists of new videos captured at additional locations. The extended dataset consists of $37$ videos with annotations provided for $711$ keyframes. The pixel-level annotations are provided for four background classes viz. greenery, construction, road and water bodies. The videos are captured at $29$ frames per second and at a resolution of $1280\times 720$ pixels. The keyframes are identified by following the shot boundary detection approach mentioned in \cite{19} and on an average, a shot consists of 15-20  frames. More details of this dataset can be found in \cite{19}.  The ManipalUAVid presented in \cite{19} contains 33 videos and annotations were provided for 667 keyframes.  Besides, the performance of semantic segmentation algorithms which analyses each keyframe individually was provided in \cite{19} on the ManipalUAVid dataset. The earlier version of ManipalUAVid dataset \cite{19} consists of last two keyframes of each video in the test split which might not be sufficient to observe the temporal smoothness or the error (if any) accumulated over the period of time in the video. Therefore, in this work, ManipalUAVid is extended by incorporating four new videos (total key frames: 44) which are entirely in the test split. Besides, the training-test split distribution is slightly modified so that a greater number of frames (4-5 frames) per video is included in the test split of this updated dataset. This aids in evaluating the video semantic segmentation models for temporal consistency.

\par Following the same protocol \cite{19}, the performance of UVid-Net is evaluated by comparing the keyframes segmented using UVid-Net with the ground truth. In ManipalUAVid, middle frames of a shot ($f^{l}_{(n/2)}$) are considered as the keyframes.  As discussed earlier, two frames ($f_{(n/2+1)}^{(l-1)}$ and $f_{(n/2)}^l$ ) are provided as the input to UVid-Net for semantic segmentation  of $f_{(n/2)}^l$ ($l \neq 1$). The dataset is divided into train, validation and test split which consists of $569$, $71$ and $71$ keyframes respectively. The following metrics are computed to evaluate the performance of the proposed method: mean Intersection over Union (mIoU), Precision, Recall and F1-score.  It may be noted that the values of the evaluation metrics obtained in this study are different from that reported in \cite{19} due to additional videos being added in the dataset.

\begin{figure}[!ht]
	
	\begin{tabular}{ccc}
		\begin{minipage}{60pt}
			\includegraphics[width=1in, height=4in]{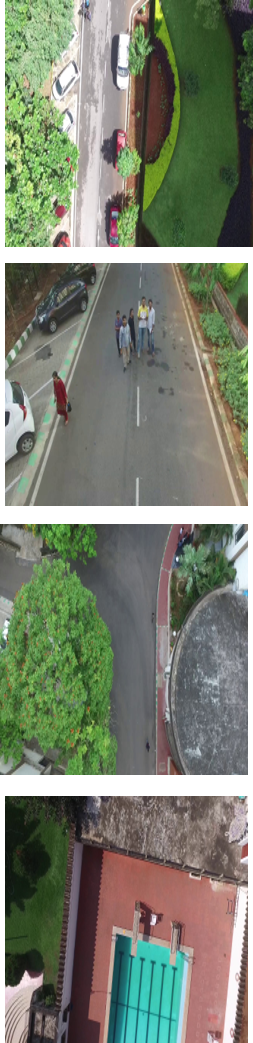}
			\centering{(a)}
		\end{minipage}
		&
		\hspace{0.15cm}
		\begin{minipage}{60pt}
			\includegraphics[width=1in, 	height=4in]{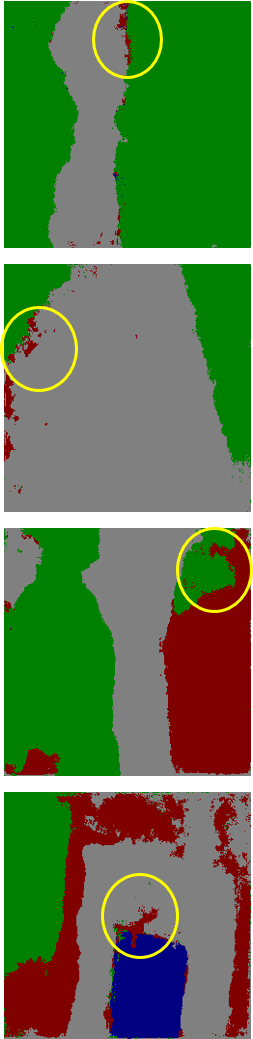}
			\centering{(b)}
		\end{minipage}
		&
		\hspace{0.15cm}
		\begin{minipage}{60pt}
			\includegraphics[width=1in, 	height=4in]{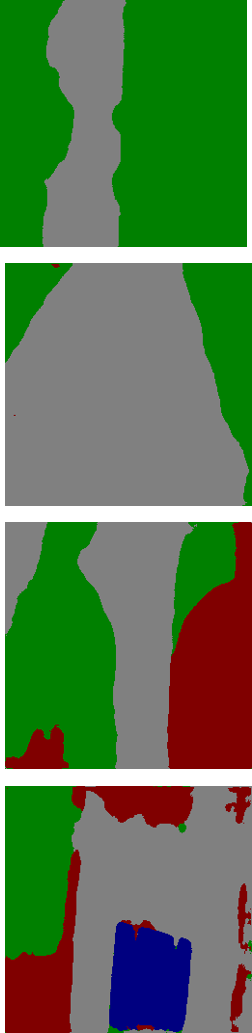}
			\centering{(c)}
		\end{minipage}
	\end{tabular}
	\caption{ Evaluating decoder: Segmentation results obtained using UVid-Net (U-Net encoder, Concatenation) and UVid-Net (U-Net encoder, Multiplication). Column (a) shows the original image, while column (b) represents the results of concatenation of feature maps (UVid-Net (Concatenation)) and the column (c) represents the results of element wise multiplication (UVid-Net (Multiplication)). Note the improvement in the segmentation by utilizing element wise multiplication (yellow circles).  Here, green, gray, red and blue  colour represents the greenery, road, construction and water bodies class respectively. }
	\label{fig:decoder}
	
\end{figure}

\begin{table*}[!h]
	\begin{center}
		\begin{tabular}{|l|c|c|c|c|c|c|}
			\hline
			Method & Precision&Recall&F1-Score &mIoU &Learnable & FLOPs\\
			&   &   &   &   &Parameters&\\
			\hline\hline
			U-Net \cite{2} &0.89&	0.89&	0.89	&	0.75&21,593,732 &62,050,187\\
			\hline
			FCN-8 \cite{4} & 0.85&	0.85&	0.85&		0.64&134,526,856& 269,028,892 \\
			\hline
			DeepLabV3+ (MobileNet-V2 backbone) \cite{33} &0.85&	0.85&	0.85&		0.65&2,142,276& 4,218,531\\
			\hline
			\hline
			TDNet \cite{hu2020temporally} & 0.85&	0.83&0.82&	0.52&	28,200,000&	6,380,000,000\\
			\hline
			Video propagation and label relaxation\cite{22}  &  0.89&0.88 &0.88 & 0.72&137,100,096&91,055,000,000\\
			\hline 
			FCN-8 + ConvLSTM \cite{wang2019deep}  &0.86 & 0.86&0.85 &0.61&134,629,100&269,821,618 \\
			\hline 
			U-Net + ConvLSTM & 0.90& 0.90& 0.90&0.76&21,695,976&62,842,913\\
			\hline
			DeepLabV3+ + ConvLSTM &0.87 &0.86 &0.85 &0.62&2,244,520&5,011,257\\
			\hline
			
			\hline
			UVid-Net (U-Net encoder)&\textbf{0.91}&	\textbf{0.91}&	\textbf{0.91}&		\textbf{0.79}&23,745,032&142,291,710\\
			\hline
			UVid-Net (ResNet-50 encoder)&0.90	&0.89	&0.89	&0.72&44,740,420&133,871,366\\
			\hline
			UVid-Net(Transfer learning) &0.89&0.88&0.87&0.60&260&-\\\hline
			
		\end{tabular}
	\end{center}
	\caption{Performance metrics of the various algorithms on ManipalUAVid dataset.}
	\label{table1}
\end{table*}

\begin{table*}[!h]
	\begin{center}
		\begin{tabular}{|l|c|c|c|c|c|}
			\hline
			Method & IoU &IoU &IoU&IoU&mIoU \\
			&(Greenery)&(Road)&(Construction)& (Water bodies)&\\
			\hline\hline
			U-Net \cite{2} &0.86	&0.81	&0.56	&0.79	&0.75	 \\
			\hline
			FCN-8 \cite{4} &0.83 &0.75&0.50	&0.48&0.64 \\
			\hline
			DeepLabV3+ \cite{33} & 0.79&0.75& 0.59&0.47	&0.65\\
			\hline
			\hline
			TDNet \cite{hu2020temporally} & 0.78	&0.72&0.51&	0.08& 0.52\\
			\hline
			Video propagation and label relaxation \cite{22}  &0.82  &0.80 &\textbf{0.67}&0.61 &0.72\\
			\hline 
			FCN-8 + ConvLSTM \cite{wang2019deep}  &0.84 &0.75 &0.49&0.39 &0.61 \\
			\hline 
			U-Net + ConvLSTM &0.87 &0.82 &0.56 &0.82&0.76\\
			\hline
			DeepLabV3+ + ConvLSTM &0.81 &0.76 &0.57 &0.28 &0.62\\
			\hline
			\hline
			UVid-Net (U-Net encoder)&0.87&	\textbf{0.86}&	0.60&\textbf{0.86}&	\textbf{0.79}\\
			\hline
			UVid-Net (ResNet-50 encoder) &0.88	&0.82	&	0.50&0.69	&0.72\\
			\hline
			UVid-Net(Transfer learning) &\textbf{ 0.89}& 0.80 &0.54&0.2&0.60\\\hline
		\end{tabular}
	\end{center}
	\caption{Per class iou and mIoU of various algorithms on ManipalUAVid dataset.}
	\label{table2}
\end{table*}

\begin{table*}[!h]
	\begin{center}
		\begin{tabular}{|l|c|c|c|c|c|c|}
			\hline
		UVid-Net & Precision&Recall&F1 &mIoU & Learnable&FLOPs \\
		Variations&&&-Score&&Parameters&\\
			\hline\hline
			U-Net encoder (Concatenation) &0.91	&0.90	&0.90	&0.78&26,878,472&161,093,886\\

			\hline
		ResNet-50 encoder (Concatenation) &	0.81 &	0.82&0..80	& 0.53 &47,801,668& 152,672,006\\
			
			\hline
			U-Net encoder (Multiplication)&\textbf{0.91}&	\textbf{0.91}&	\textbf{0.91}&		\textbf{0.79}&23,745,032&142,291,710\\
			
			\hline
			ResNet-50 encoder (Multiplication)&0.90	&0.89	&0.89	&0.72&44,740,420&133,871,366\\
			
			\hline
	
		\end{tabular}
	\end{center}
	\caption{Comparing performance of UVid-Net (Concatenation) with UVid-Net (Multiplication).}
	\label{table3}
\end{table*}
\begin{table}[!h]
	\begin{center}
		\begin{tabular}{|l|c|c|c|c|c|}
			\hline
			UVid-Net & IoU &IoU &IoU&IoU&mIoU \\
			Variations&(Greenery)&(Road)&(Constr-& (Water &\\
			&          &      & uction)              &bodies)&\\
			\hline
			\hline
	    	U-Net encoder&0.88 &	0.83 &	0.60 &	0.84 &0.78\\
	    	(Concatenation)&&&&&\\
			\hline
	    	ResNet-50 encoder&0.84 &	0.72 &	0.17 &	0.40 &0.53\\
	    	(Concatenation)&&&&&\\
			\hline
			U-Net encoder& 0.87 &	\textbf{0.86} &	0.60 &\textbf{0.86} &	\textbf{0.79}\\
			(Multiplication)&&&&&\\
			\hline
		
		ResNet-50 encoder&0.88	&0.82	&	0.50&0.69	&0.72\\
		(Multiplication)&&&&&\\
		\hline
		\end{tabular}
	\end{center}
	\caption{Per class iou and mIoU of UVid-Net for comparing performance of UVid-Net (Concatenation) with UVid-Net (Multliplication).}
	\label{table4}
\end{table}

\begin{figure}[h]
	\begin{center}
		\includegraphics[width=\columnwidth,height=3.3in]{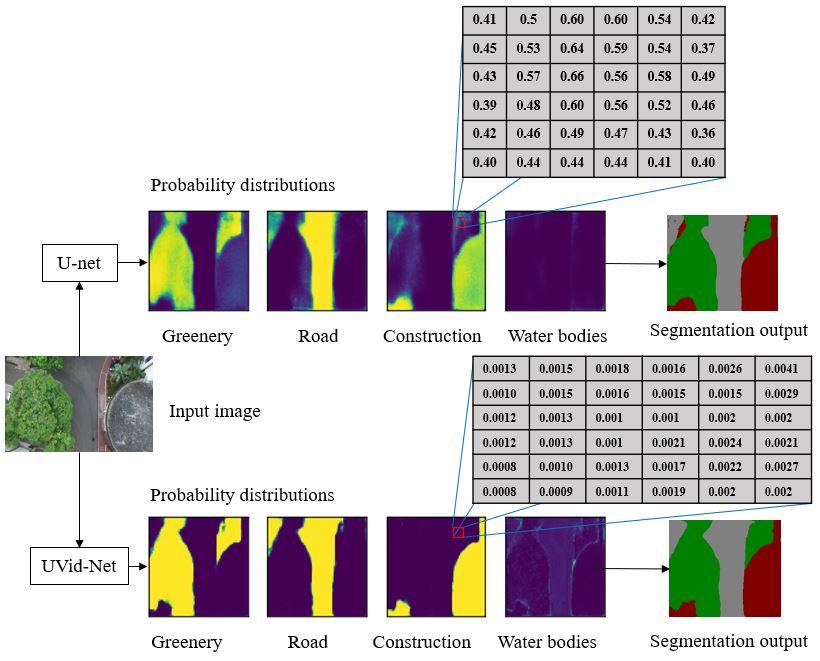}
	\end{center}
	\caption{The heat map of probability distributions produced by U-Net and UVid-Net algorithm. Row 1 shows the softmax output of U-Net while row 2 shows the softmax output of UVid-Net. Also, shown are the actual softmax output for a 6 x 6 region.}
	\label{fig:encoder}
\end{figure}

\subsection{Evaluation of encoder}
\label{Sec:TempInfo}

The proposed encoder part consists of two branches which extract features from two consecutive keyframes of a video simultaneously.  Two variants of UVid-Net (U-Net encoder and ResNet-50 encoder) encoders are considered in this work. To evaluate the performance of the proposed architecture, we compare it with the traditional U-Net architecture (with a single encoder branch). Figure \ref{fig:temporal} shows the comparison of the segmentation results obtained using a single branch U-Net and two branch UVid-Net. Since single branch U-Net is an image semantic segmentation algorithm, it fails to capture the temporal information and hence produces temporally inconsistent labels. In contrast, the proposed architecture is able to capture the temporal dynamics between the two keyframes and produces more accurate results. For instance, the U-Net with single branch encoder incorrectly classifies few pixels belonging to road/greenery class as construction (shown in yellow circles).   However, the two branch encoder based proposed method correctly classifies these pixels as road/greenery, thus producing a temporally smoother segmentation result as shown in Figure \ref{fig:temporal} (d) and Figure \ref{fig:temporal} (e).

Tables \ref{table1} and Table  \ref{table2} compares the performance of single branch encoder U-Net with the proposed UVid-Net in terms of mIoU, Precision, Recall and F1-score. It is observed that the per class IoU of UVid-Net (U-Net encoder) for all the four classes are higher than the single branch U-Net. Moreover, from Table \ref{table1} it is observed the proposed method has higher recall and precision scores than single branch U-Net which indicates that it has produced lower false positives and false negatives. The above results demonstrate the effectiveness of two branch encoder module in acquiring temporal information and thus resulting in a more accurate segmentation as compared to the classical single branch encoder U-Net. It may be noted that a single branch U-Net with ResNet-50 encoder suffered from high variance (over-fitting), even in the presence of regularization, with a training and validation accuracy of 0.98 and 0.66 respectively.  

Table \ref{table1} and \ref{table2} also compare the performance of encoder architectures based on U-Net encoder and ResNet-50. It can be observed that the proposed encoder module based on U-Net encoder and ResNet-50 achieves a comparable performance (in terms of IoU, Table \ref{table2}) on greenery and road class while a slightly lower IoU is observed construction and waterbodies class for ResNet-50 based encoder. This decrease in IoU for waterbodies and construction class for ResNet-50 based encoder is on expected lines due to the challenges encountered in learning the parameters of a deeper network (ResNet-50) with limited training images. The decrease in IoU for two classes results in a slightly lower mIoU for UVid-Net based on ResNet-50 encoder as compared to that of U-Net encoder. However, in spite of the decrease in IoU for two classes, the overall mIoU obtained using UVid-Net with ResNet-50 based encoder (0.72) is comparable with that of the current state-of-the-art method \cite{22}.

In addition to the qualitative and quantitative evaluation of the encoder, the softmax output of U-Net and UVid-Net (U-Net encoder) is also analysed in Figure \ref{fig:encoder}. It can be observed that a high probability score is obtained for the pixels in their actual class in UVid-Net as compared to that of U-Net. The high probability score eliminates uncertainty and produces a more accurate segmentation. For example, a high probability score for greenery class is obtained for pixel belonging to trees using UVid-Net (Figure \ref{fig:encoder}). In addition, U-Net which lacks temporal information has produced higher construction class probability for pixel belonging to greenery at the boundaries (Refer the 6 $\times$ 6 representative regions in Figure \ref{fig:encoder}). In contrast, the UVid-Net which utilizes features propagated from the previous frame has produced very low construction class probability for greenery pixels at the class boundaries.

\subsection{Evaluation of decoder}
\label{Sec:Decoder}
The decoder of the proposed UVid-Net architecture consists of skip connections from the lower branch of the encoder to the corresponding decoder layers. Element-wise multiplication operation is utilized to combine the activations of the encoder and decoder layers. The experimental evaluation of the proposed feature-refiner module with the concatenation approach suggests a marginal increase in mIoU for UVid-Net with U-Net encoder ( Tables \ref{table3} and \ref{table4} ).  It can be observed that the per class IoU is higher for road and water bodies for the multiplication operation as compared to concatenation. Further, the other two classes (greenery and construction) performs competitively in terms of per class IoU. However, the qualitative evaluation shows that a more accurate segmentation is obtained using the proposed approach compared with the concatenation. Figure \ref{fig:decoder} shows few images where finer segmentation boundaries are obtained using the UVid-Net (multiplication) with U-Net encoder as compared to UVid-Net (concatenation). It may be observed in Figure \ref{fig:decoder} (First two rows), that the pixels from the road class have been misclassified as construction class using UVid-Net (concatenation), while a precise greenery-road boundary is obtained using UVid-Net (multiplication).  The improvement obtained using the proposed feature-refiner module is more prominent for UVid-Net (ResNet encoder).  A mIoU of 0.72 is obtained with UVid-Net (ResNet encoder) with multiplication operation as compared to 0.53 with concatenation.

Moreover, the feature-refiner module reduces the number of FLOPs along with a number of parameters. It is observed that UVid-Net (multiplication) results in 142,291,716 FLOPs while UVid-Net (concatenation) results in 161,093,892 FLOPs for U-Net encoder ($\sim$ 11\% less FLOPs). These results show that an accurate segmentation is obtained using UVid-Net (multiplication) with much less computation overhead. Besides, the element-wise multiplication operation in UVid-Net also reduces the number of learnable parameters  ($23,745,032$) in the network as compared to the concatenation in the UVid-Net ($26,862,856$). This result is significant since the proposed architecture produces higher mIoU (in the order of 0.79) with a reduced number of parameters. Indeed, the reduced complexity and the number of parameters of UVid-Net as compared to traditional concatenation operation makes it an ideal CNN architecture which can be used for UAV-based IoT applications.

\begin{figure}[!ht]
	
	\begin{tabular}{cccc}
		\begin{minipage}{32pt}
			\includegraphics[width=0.8in, height=0.8in]{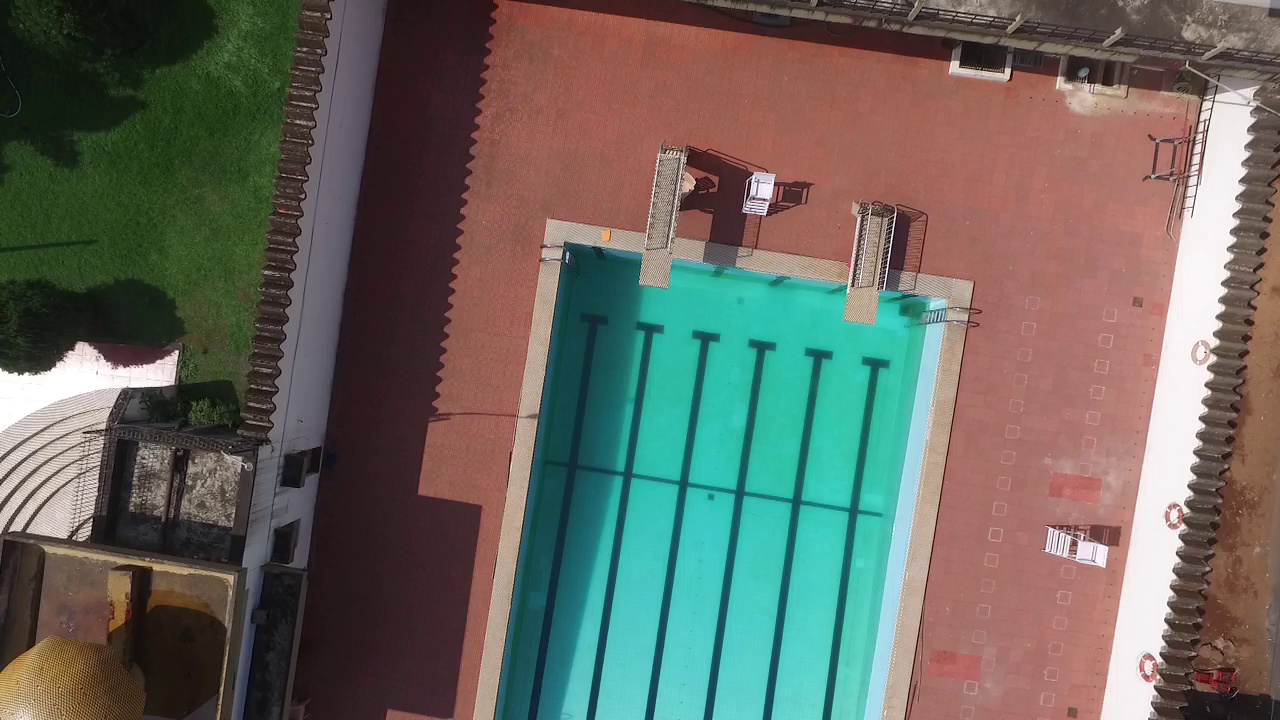}
		\end{minipage}
		&
		\hspace{0.5cm}
		\begin{minipage}{32pt}
			\includegraphics[width=0.8in, 	height=0.8in]{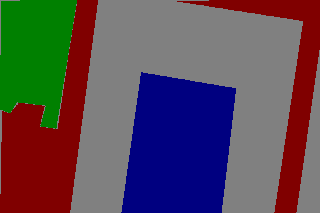}
		\end{minipage}
		&
		\hspace{0.5cm}
		\begin{minipage}{32pt}
			\includegraphics[width=0.8in, 	height=0.8in]{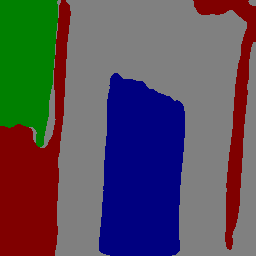}
		\end{minipage}
		&
		\hspace{0.5cm}
		\begin{minipage}{32pt}
			\includegraphics[width=0.8in, 	height=0.8in]{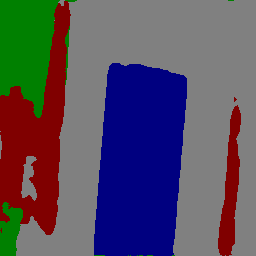}
		\end{minipage}
		\\
		\\
		\begin{minipage}{32pt}
			\includegraphics[width=0.8in, height=0.8in]{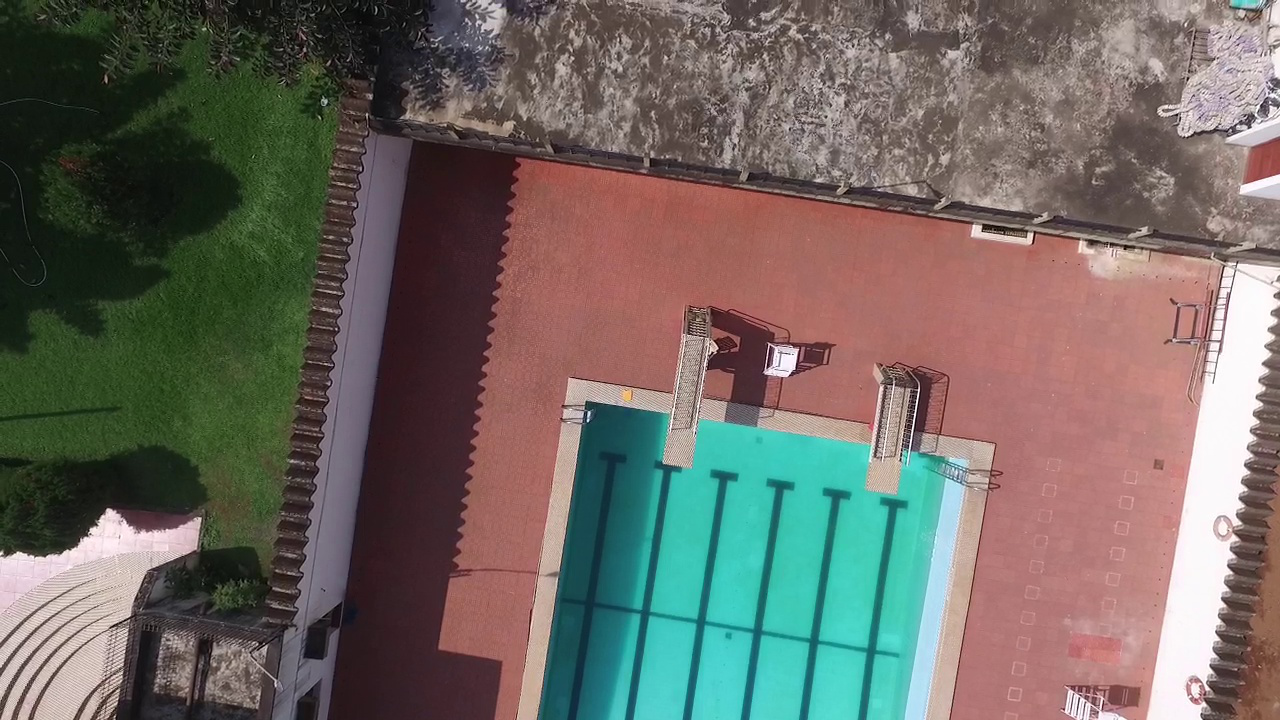}
			\centering{(a)}
		\end{minipage}
		&
		\hspace{0.5cm}
		\begin{minipage}{32pt}
			\includegraphics[width=0.8in, 	height=0.8in]{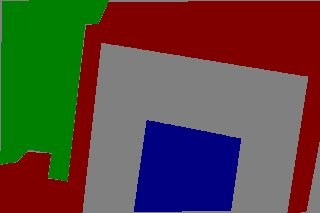}
			\centering{(b)}
		\end{minipage}
		&
		\hspace{0.5cm}
		\begin{minipage}{32pt}
			\includegraphics[width=0.8in, 	height=0.8in]{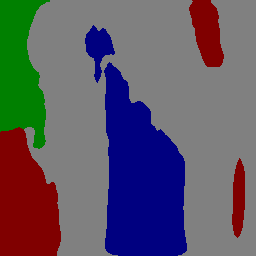}
			\centering{(c)}
		\end{minipage}
		&
		\hspace{0.5cm}
		\begin{minipage}{32pt}
			\includegraphics[width=0.8in, 	height=0.8in]{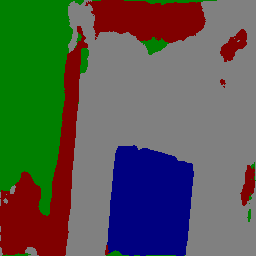}
			\centering{(d)}
		\end{minipage}
		\\
		\\
		
	\end{tabular}
	\caption{ Comparing performance of U-Net + ConvLSTM with UVid-Net (U-Net encoder). Column (a) and (b) shows two consecutive key frames and its corresponding ground-truth. Column (c) and (d) shows the results of U-Net with ConvLSTM and UVid-Net. Here, green, gray, red and blue  colour represents the greenery, road, construction and water bodies class respectively. }
	\label{fig:convu}
	
\end{figure}

\begin{figure*}[!t]
	
	\begin{tabular}{cccccccc}
	    \begin{minipage}{32pt}
			\includegraphics[width=0.74in, height=0.74in]{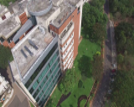}
		\end{minipage}
		&
		\hspace{0.4cm}
		\begin{minipage}{32pt}
			\includegraphics[width=0.74in, height=0.74in]{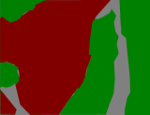}
		\end{minipage}
		&
		\hspace{0.4cm}
		\begin{minipage}{32pt}
			\includegraphics[width=0.74in, height=0.74in]{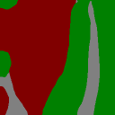}
		\end{minipage}
		&
		\hspace{0.4cm}
		\begin{minipage}{32pt}
			\includegraphics[width=0.74in, height=0.74in]{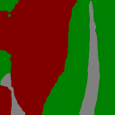}
		\end{minipage}
		&
		\hspace{0.4cm}
		\begin{minipage}{32pt}
			\includegraphics[width=0.74in, height=0.74in]{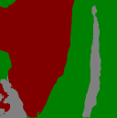}
		\end{minipage}
		&
		
		\hspace{0.4cm}
		\begin{minipage}{32pt}
			\includegraphics[width=0.74in, height=0.74in]{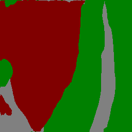}
		\end{minipage}
		&
		\hspace{0.4cm}
		\begin{minipage}{32pt}
			\includegraphics[width=0.74in, height=0.74in]{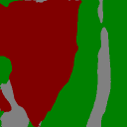}
		\end{minipage}
		&
		\hspace{0.4cm}
		\begin{minipage}{32pt}
			\includegraphics[width=0.74in, height=0.74in]{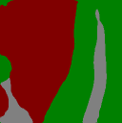}
		\end{minipage}
		\\
		\\
		\begin{minipage}{32pt}
			\includegraphics[width=0.74in, height=0.74in]{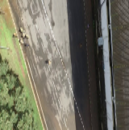}
		\end{minipage}
		&
		\hspace{0.4cm}
		\begin{minipage}{32pt}
			\includegraphics[width=0.74in, height=0.74in]{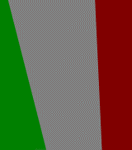}
		\end{minipage}
		&
		\hspace{0.4cm}
		\begin{minipage}{30pt}
			\includegraphics[width=0.74in, height=0.74in]{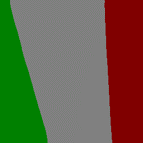}
		\end{minipage}
		&
		\hspace{0.4cm}
		\begin{minipage}{30pt}
			\includegraphics[width=0.74in, height=0.74in]{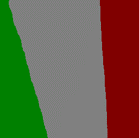}
		\end{minipage}
		&
		\hspace{0.4cm}
		\begin{minipage}{30pt}
			\includegraphics[width=0.74in, height=0.74in]{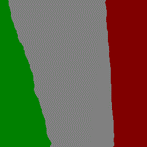}
		\end{minipage}
		&
	
		\hspace{0.4cm}
		\begin{minipage}{30pt}
			\includegraphics[width=0.74in, height=0.74in]{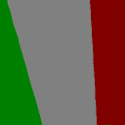}
		\end{minipage}
		&
		\hspace{0.4cm}
		\begin{minipage}{30pt}
			\includegraphics[width=0.74in, height=0.74in]{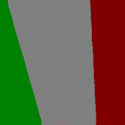}
		\end{minipage}
		&
		\hspace{0.4cm}
		\begin{minipage}{30pt}
			\includegraphics[width=0.74in, height=0.74in]{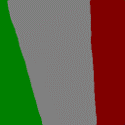}
		\end{minipage}
		\\
		\\
		\begin{minipage}{30pt}
			\includegraphics[width=0.74in, height=0.74in]{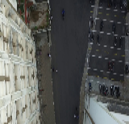}
		\end{minipage}
		&
		\hspace{0.4cm}
		\begin{minipage}{30pt}
			\includegraphics[width=0.74in, height=0.74in]{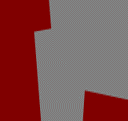}
		\end{minipage}
		&
		\hspace{0.4cm}
		\begin{minipage}{30pt}
			\includegraphics[width=0.74in, height=0.74in]{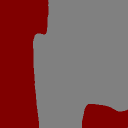}
		\end{minipage}
		&
		\hspace{0.4cm}
		\begin{minipage}{30pt}
			\includegraphics[width=0.74in, height=0.74in]{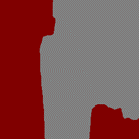}
		\end{minipage}
		&
		\hspace{0.4cm}
		\begin{minipage}{30pt}
			\includegraphics[width=0.74in, height=0.74in]{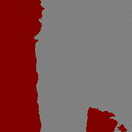}
		\end{minipage}
		&
	
		\hspace{0.4cm}
		\begin{minipage}{30pt}
			\includegraphics[width=0.74in, height=0.74in]{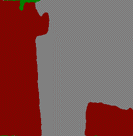}
		\end{minipage}
		&
		\hspace{0.4cm}
		\begin{minipage}{30pt}
			\includegraphics[width=0.74in, height=0.74in]{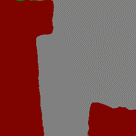}
		\end{minipage}
		&
		\hspace{0.4cm}
		\begin{minipage}{30pt}
			\includegraphics[width=0.74in, height=0.74in]{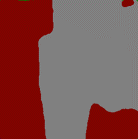}
		\end{minipage}
		\\
		\\
		\begin{minipage}{30pt}
			\includegraphics[width=0.74in, height=0.74in]{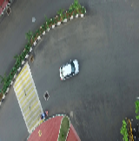}
		\end{minipage}
		&
		\hspace{0.4cm}
		\begin{minipage}{30pt}
			\includegraphics[width=0.74in, height=0.74in]{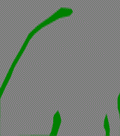}
		\end{minipage}
		&
		\hspace{0.4cm}
		\begin{minipage}{30pt}
			\includegraphics[width=0.74in, height=0.74in]{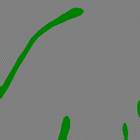}
		\end{minipage}
		&
		\hspace{0.4cm}
		\begin{minipage}{30pt}
			\includegraphics[width=0.74in, height=0.74in]{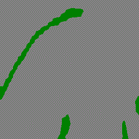}
		\end{minipage}
		&
		\hspace{0.4cm}
		\begin{minipage}{30pt}
			\includegraphics[width=0.74in, height=0.74in]{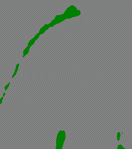}
		\end{minipage}
		
		&
		\hspace{0.4cm}
		\begin{minipage}{30pt}
			\includegraphics[width=0.74in, height=0.74in]{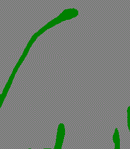}
		\end{minipage}
		&
		\hspace{0.4cm}
		\begin{minipage}{30pt}
			\includegraphics[width=0.74in, height=0.74in]{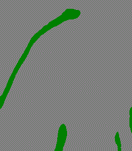}
		\end{minipage}
		&
		\hspace{0.4cm}
		\begin{minipage}{30pt}
			\includegraphics[width=0.74in, height=0.74in]{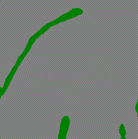}
		\end{minipage}
		\\
		\\
		\begin{minipage}{30pt}
			\includegraphics[width=0.74in, height=0.74in]{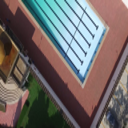}
		\end{minipage}
		&
		\hspace{0.4cm}
		\begin{minipage}{30pt}
			\includegraphics[width=0.74in, height=0.74in]{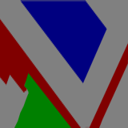}
		\end{minipage}
		&
		\hspace{0.4cm}
		\begin{minipage}{30pt}
			\includegraphics[width=0.74in, height=0.74in]{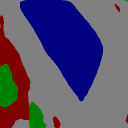}
		\end{minipage}
		&
		\hspace{0.4cm}
		\begin{minipage}{30pt}
			\includegraphics[width=0.74in, height=0.74in]{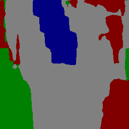}
		\end{minipage}
		&
		\hspace{0.4cm}
		\begin{minipage}{30pt}
			\includegraphics[width=0.74in, height=0.74in]{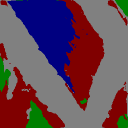}
		\end{minipage}
		&
		
		\hspace{0.4cm}
		\begin{minipage}{30pt}
			\includegraphics[width=0.74in, height=0.74in]{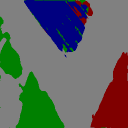}
		\end{minipage}
		&
		\hspace{0.4cm}
		\begin{minipage}{30pt}
			\includegraphics[width=0.74in, height=0.74in]{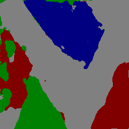}
		\end{minipage}
		&
		\hspace{0.4cm}
		\begin{minipage}{30pt}
			\includegraphics[width=0.74in, height=0.74in]{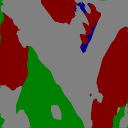}
		\end{minipage}
		\\
		\\
		\begin{minipage}{30pt}
			\includegraphics[width=0.74in, height=0.74in]{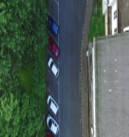}
		\end{minipage}
		&
		\hspace{0.4cm}
		\begin{minipage}{30pt}
			\includegraphics[width=0.74in, height=0.74in]{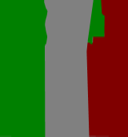}
		\end{minipage}
		&
		\hspace{0.4cm}
		\begin{minipage}{30pt}
			\includegraphics[width=0.74in, height=0.74in]{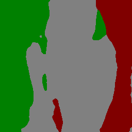}
		\end{minipage}
		&
		\hspace{0.4cm}
		\begin{minipage}{30pt}
			\includegraphics[width=0.74in, height=0.74in]{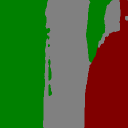}
		\end{minipage}
		&
		\hspace{0.4cm}
		\begin{minipage}{30pt}
			\includegraphics[width=0.74in, height=0.74in]{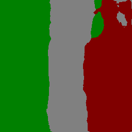}
		\end{minipage}
	
		&
		\hspace{0.4cm}
		\begin{minipage}{30pt}
			\includegraphics[width=0.74in, height=0.74in]{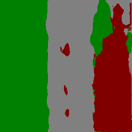}
		\end{minipage}
		&
		\hspace{0.4cm}
		\begin{minipage}{30pt}
			\includegraphics[width=0.74in, height=0.74in]{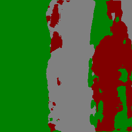}
		\end{minipage}
		&
		\hspace{0.4cm}
		\begin{minipage}{30pt}
			\includegraphics[width=0.74in, height=0.74in]{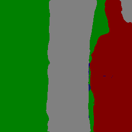}
		\end{minipage}
		\\
		\\
		\begin{minipage}{30pt}
			\includegraphics[width=0.74in, height=0.74in]{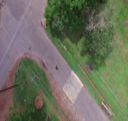}
		\end{minipage}
		&
		\hspace{0.4cm}
		\begin{minipage}{30pt}
			\includegraphics[width=0.74in, height=0.74in]{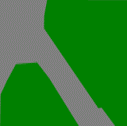}
		\end{minipage}
		&
		\hspace{0.4cm}
		\begin{minipage}{30pt}
			\includegraphics[width=0.74in, height=0.74in]{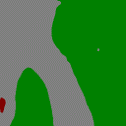}
		\end{minipage}
		&
		\hspace{0.4cm}
		\begin{minipage}{30pt}
			\includegraphics[width=0.74in, height=0.74in]{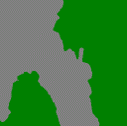}
		\end{minipage}
		&
		\hspace{0.4cm}
		\begin{minipage}{30pt}
			\includegraphics[width=0.74in, height=0.74in]{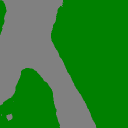}
		\end{minipage}
	
		&
		\hspace{0.4cm}
		\begin{minipage}{30pt}
			\includegraphics[width=0.74in, height=0.74in]{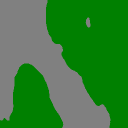}
		\end{minipage}
		&
		\hspace{0.4cm}
		\begin{minipage}{30pt}
			\includegraphics[width=0.74in, height=0.74in]{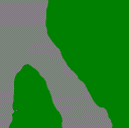}
		\end{minipage}
		&
		\hspace{0.4cm}
		\begin{minipage}{30pt}
			\includegraphics[width=0.74in, height=0.74in]{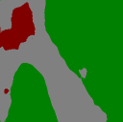}
		\end{minipage}
		\\
		\\
		\begin{minipage}{30pt}
			\includegraphics[width=0.74in, height=0.74in]{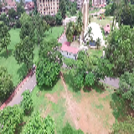}
			\centering{(a)}
		\end{minipage}
		&
		\hspace{0.4cm}
		\begin{minipage}{30pt}
			\includegraphics[width=0.74in, height=0.74in]{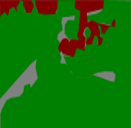}
			\centering{(b)}
		\end{minipage}
		&
		\hspace{0.4cm}
		\begin{minipage}{30pt}
			\includegraphics[width=0.74in, height=0.74in]{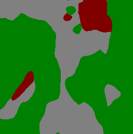}
			\centering{(c)}
		\end{minipage}
		&
		\hspace{0.4cm}
		\begin{minipage}{30pt}
			\includegraphics[width=0.74in, height=0.74in]{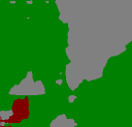}
			\centering{(d)}
		\end{minipage}
		&
		\hspace{0.4cm}
		\begin{minipage}{30pt}
			\includegraphics[width=0.74in, height=0.74in]{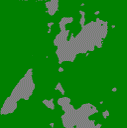}
			\centering{(e)}
		\end{minipage}
		
		&
		\hspace{0.4cm}
		\begin{minipage}{30pt}
			\includegraphics[width=0.74in, height=0.74in]{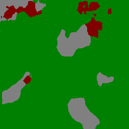}
			\centering{(f)}
		\end{minipage}
		&
		\hspace{0.4cm}
		\begin{minipage}{30pt}
			\includegraphics[width=0.74in, height=0.74in]{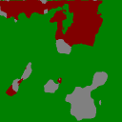}
			\centering{(g)}
		\end{minipage}
		&
		\hspace{0.4cm}
		\begin{minipage}{30pt}
			\includegraphics[width=0.74in, height=0.74in]{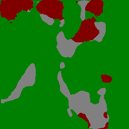}
			\centering{(h)}
		\end{minipage}
		\\
		\\
			\begin{minipage}{30pt}
			\includegraphics[width=0.74in, height=0.3in]{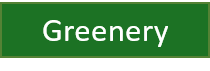}
		\end{minipage}
		&
		\hspace{0.4cm}
		\begin{minipage}{30pt}
			\includegraphics[width=0.74in, height=0.3in]{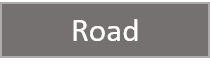}
		\end{minipage}
		&
		\hspace{0.4cm}
		\begin{minipage}{30pt}
			\includegraphics[width=0.74in, height=0.3in]{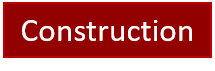}
		\end{minipage}
		&
		\hspace{0.4cm}
		\begin{minipage}{30pt}
			\includegraphics[width=0.74in, height=0.3in]{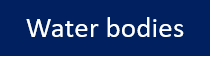}
		\end{minipage}
		&
		\hspace{0.4cm}
		\begin{minipage}{30pt}
		\end{minipage}
		&
		\hspace{0.4cm}
		\begin{minipage}{30pt}
		\end{minipage}
		&
		\hspace{0.4cm}
		\begin{minipage}{30pt}
		\end{minipage}
		&
	
		\hspace{0.4cm}
		\begin{minipage}{30pt}
		\end{minipage}
		\\
		\\

	\end{tabular}
	\caption{ UAV Video Semantic Segmentation Results on ManipalUAVid dataset \cite{19}:  Column (a) and (b) shows the keyframes from UAV video and its corresponding ground truth. Column (c) and (d) shows the results of ConvLSTM with U-Net and FCN8 backbone architectures respectively. Column (e) shows the results of \cite{22}. Column (f) shows the results of UVid-Net with ResNet50 encoder and column (g) shows the results of UVid-Net with U-Net encoder. Column (h) shows the results of transfer learning.}
	\label{fig:SOTA}
	
\end{figure*}

\subsection {Comparison with state-of-art}
\label{Sec:SOTAComp}

\par The proposed approach is compared with the existing state-of-the-art image semantic segmentation methods viz. U-Net \cite{2},  FCN8 \cite{4} and DeepLabV3+ \cite{33}. However, these methods do not incorporate temporal information and segments each keyframe independently. Therefore, the proposed method is also compared with the state-of-the-art approaches (\cite{22} and \cite{hu2020temporally}) on CityScape dataset that include temporal information. The authors in \cite{hu2020temporally} used multiple shallow CNNs to extract features from multiple frames. Subsequently, attention mechanism is utilized to combine the temporal features. In \cite{22}, the authors proposed to use video prediction model to propagate labels to the immediate neighbouring frames for creating more image-label pairs \cite{22}.  Besides, the performance of UVidNet is compared with the UAV \textit{video} semantic segmentation approach proposed in \cite{wang2019deep}. This approach uses a Convolution Long Short Term Memory (Conv-LSTM) module to capture the temporal dynamics of the video. It may be noted that the method proposed in  \cite{wang2019deep} independently segments each frames using FCN8, and then the resulting frames are passed through Conv-LSTM module as the post-processing step. However, in addition to combining FCN8 + Conv-LSTM, we also compare the performance by segmenting individual frames with U-Net/DeepLabV3+ and then post-processing it with Conv-LSTM module, resulting in two additional methods viz UNet + ConvLSTM and DeepLabV3+ + ConvLSTM.

The proposed architecture is quantitatively compared with the above mentioned existing approaches. Table \ref{table1} compares the performance metrics such as precision, recall, F1-score and mean Intersection over Union (mIoU) while Table \ref{table2} compares the per class IoU and mIoU of the existing methods with the proposed method. As discussed earlier, the image semantic segmentation approaches (UNet, FCN, DeepLabV3+) segments each keyframe independently and fails to capture temporal cues. It can be observed a mIoU of 0.79 is obtained by the proposed approach as compared to a mIoU of 0.75, 0.64 and 0.65 for UNet, FCN8 and DeepLabV3+ respectively. The proposed approach outperforms the existing image segmentation approach. Besides, it can be observed from Figure \ref{fig:SOTA} that UVid-Net produces a more accurate segmentation map with smoother segmentation boundaries as compared with other approaches. The proposed UVid-Net incorporates temporal information by merging the features extracted from two different frames of a video and thereby outperforms the existing image semantic segmentation algorithms.

In addition to the image segmentation algorithms, the proposed approach is also compared with the video semantic segmentation algorithms viz. TDNet \cite{hu2020temporally}, Video Propagation /Label Relaxation \cite{22}, UNet-ConvLSTM, FCN8-ConvLSTM \cite{wang2019deep}, and DeepLabV3+ - ConvLSTM. It can be seen (Table \ref{table1})  that the UVid-Net (U-Net encoder) achieves a mIoU of 0.79 and F1-score of 0.91  outperforming the other video segmentation approaches. Besides, UVid-Net (ResNet50-encoder) performs competitively and achieves F1-score of 0.89 and a mIoU of 0.72. To study the performance of the proposed method for each class, the per-class IoU is computed as shown in Table \ref{table2}.

\begin{figure}[!ht]
	
	\begin{tabular}{cccc}
		\begin{minipage}{32pt}
			\includegraphics[width=0.8in, height=0.8in]{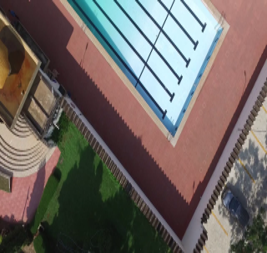}
		\end{minipage}
		&
		\hspace{0.5cm}
		\begin{minipage}{32pt}
			\includegraphics[width=0.8in, 	height=0.8in]{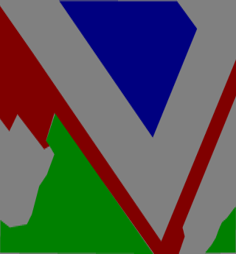}
		\end{minipage}
		&
		\hspace{0.5cm}
		\begin{minipage}{32pt}
			\includegraphics[width=0.8in, 	height=0.8in]{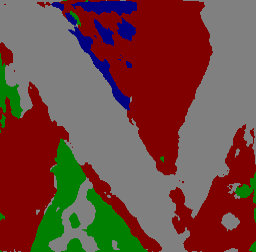}
		\end{minipage}
		&
		\hspace{0.5cm}
		\begin{minipage}{32pt}
			\includegraphics[width=0.8in, 	height=0.8in]{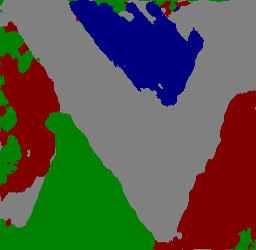}
		\end{minipage}
		\\
		\\
		\begin{minipage}{32pt}
			\includegraphics[width=0.8in, height=0.8in]{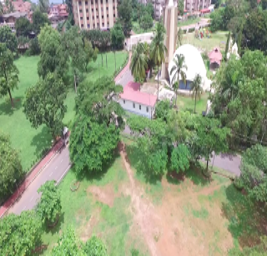}
		\end{minipage}
		&
		\hspace{0.5cm}
		\begin{minipage}{32pt}
			\includegraphics[width=0.8in, 	height=0.8in]{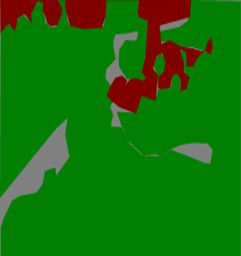}
		\end{minipage}
		&
		\hspace{0.5cm}
		\begin{minipage}{32pt}
			\includegraphics[width=0.8in, 	height=0.8in]{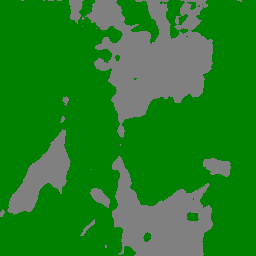}
		\end{minipage}
		&
		\hspace{0.5cm}
		\begin{minipage}{32pt}
			\includegraphics[width=0.8in, 	height=0.8in]{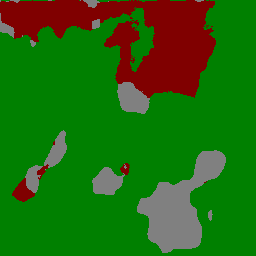}
		\end{minipage}
		\\
		\\
			\begin{minipage}{32pt}
			\includegraphics[width=0.8in, height=0.8in]{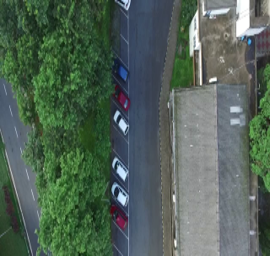}
			\centering{(a)}
		\end{minipage}
		&
		\hspace{0.5cm}
		\begin{minipage}{32pt}
			\includegraphics[width=0.8in, 	height=0.8in]{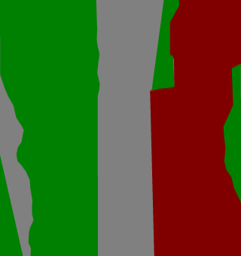}
			\centering{(b)}
		\end{minipage}
		&
		\hspace{0.5cm}
		\begin{minipage}{32pt}
			\includegraphics[width=0.8in, 	height=0.8in]{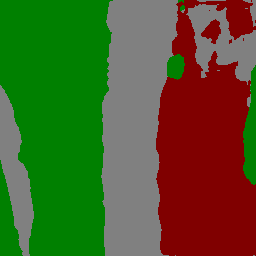}
			\centering{(c)}
		\end{minipage}
		&
		\hspace{0.5cm}
		\begin{minipage}{32pt}
			\includegraphics[width=0.8in, 	height=0.8in]{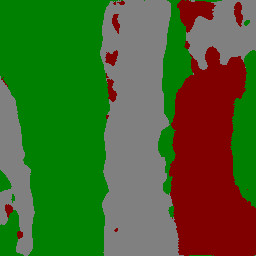}
			\centering{(d)}
		\end{minipage}
		\\
		\\
		
	\end{tabular}
	\caption{ A visual results comparison of UVid-Net and \cite{22}  on construction class. Column (a) Shows the raw image, column (b) shows the ground truth, column (c) shows the results of \cite{22} and column (d) shows the results of UVid-Net. }
	\label{fig:lp}
\end{figure}

The water bodies class account for only 1.2\% of the total pixels in the dataset. In spite of the limited annotation available, the proposed approach UVid-Net outperforms the existing methods and the current state-of-the-art \cite{22} by a significant margin (IoU of 0.86 for UVid-Net Vs 0.61 for \cite{22}). Moreover, the construction class accounts for a slightly higher pixel count (5.5\%) in the dataset. For the construction class, the proposed method outperforms the existing algorithms (except \cite{22}) in terms of IoU. A slightly higher IoU is observed using \cite{22} (0.67) as compared to UVid-Net (0.60) for the construction class. Figure \ref{fig:lp} compares the performance of \cite{22} and UVid-Net for segmentation of construction class. More accurate segmentation is obtained for few frames using the current state-of-the-art \cite{22} method for construction class as compared to the proposed method (Figure \ref{fig:lp}, row 3). However, for other frames, a more accurate segmentation is obtained for the construction class using the proposed method (Figure \ref{fig:lp}, row 1 and 2). The proposed approach performs competitively (for construction class) with the current state-of-the-art \cite{22} with a significant reduction in the model parameters and without the need for an extra sequential model/optical flow. It may also be noted that \cite{22} contains 137M parameters, \cite{hu2020temporally} contains  28M parameters, while the proposed approach contains 23M parameters.  
Besides, there is a reduction in the computational complexity (91,055,000,000 FLOPs for \cite{22}, 6,380,000,000 FLOPs for \cite{hu2020temporally}, while only 142,291,710 FLOPs for UVid-Net). Hence, the proposed method for video semantic segmentation is efficient in terms of computational complexity and is a viable solution for edge computing based applications such as scene parsing using UAV. Figure \ref{fig:SOTA} compares the segmentation results obtained using the proposed approach and the existing methods. It can be observed that the more accurate segmentation is obtained using the proposed method as compared to the existing methods. For instance, the proposed method is able to accurately identify construction, greenery and water bodies especially in fifth and eight rows of Figure \ref{fig:SOTA}.

The UNet-ConvLSTM performs competitively on ManipalUAVid dataset with a mIoU of 0.76. 
However, U-Net-ConvLSTM fails to capture the temporal dynamics as shown in Figure \ref{fig:convu}.  In comparison, UVid-Net (U-Net encoder) produces a more accurate segmentation, especially for the water body class.


In addition to the significant improvement in the performance, the UVid-Net (U-Net encoder) has a lower number of parameters as compared to the FCN-8, FCN-8 + ConvLSTM as shown in Table \ref{table1}. Further, UVid-Net (U-Net encoder) has a comparable number of parameters with other models with an exception of DeepLabV3+ which uses MobileNet-V2 backbone. The lower parameters of UVid-Net reduces the dependency on the availability of huge training data.

\subsection{Evaluation of transfer learning}
\par The availability of manually annotated training dataset of sufficient size is a challenge in supervised deep learning based approach. A widely used approach in this scenario is to train the CNN network on a huge dataset and then transfer the weights learned for the task at hand \cite{yosinski2014transferable}.


\par In this work, the transfer learning approach has been studied on UVid-Net (U-Net encoder) for semantic segmentation of UAV aerial videos. The UAVid-Net (U-Net encoder) is initially trained on Cityscape \cite{cordts2016cityscapes} dataset to predict eight categorical classes (flat, human, vehicle, construction, object, nature, sky and void) by using Adam optimizer with a learning rate set to $0.0001$. This dataset is selected due to its similarity in classes as compared to ManipalUAVid. Moreover, this dataset consists of 3000 training images which are greater than the ManipalUAVid dataset and helps in learning more generalized features. Subsequently, the last layer of the model is re-trained (with other layers frozen) on the ManipalUAVid dataset to predict four classes (greenery, road, construction and water bodies).  The performance metrics of UVid-Net (U-Net encoder) by utilizing transfer learning is shown in Table \ref{table1} and \ref{table2}. It is observed that the UVid-Net has performed competitively on greenery, road and construction classes with a per class IoU of $0.89$, $0.80$ and $0.54$ respectively. However, a low per class IoU is observed on water bodies class ($0.20$). This result was expected since the Cityscape dataset does not contain any images with water, and has no definition for water bodies class.  Figure \ref{fig:SOTA} shows the segmentation result of transfer learning on UVid-Net (U-Net encoder). It can be observed that the transfer learning approach offers competitive results as compared to existing approaches on greenery, road and construction classes.  Despite the limitation on unknown classes, pre-trained UVid-Net (U-Net encoder) could be the preferred choice especially in the case of limited availability of training dataset for UAV aerial videos segmentation.

\section{Conclusion}
\par This paper presents a new encoder-decoder based CNN architecture for semantic segmentation of UAV aerial videos. The proposed architecture utilizes a new encoder consisting of two parallel encoding branches with two consecutive keyframes of the video as the input to the network. By integrating the features extracted from the two encoding branches, the network can learn temporal information eliminating the need for an extra sequential module. Besides, it uses a feature-refiner module in the decoder path. This module produces smoother segmentation boundaries. The proposed architecture achieved a mIoU of $0.79$ on ManipalUAVid dataset which outperforms the other state-of-the-art algorithms. 
This work also demonstrated that the proposed network UVid-Net trained on a larger semantic segmentation dataset for Urban street scenes (Cityscape) can be utilized for UAV aerial videos segmentation. This transfer learning approach shows that competitive results are obtained on ManipalUAVid dataset by re-training only the last layer of UVid-Net trained on Cityscape dataset.  These results hold significance as it reduces the dependency on the availability of manually annotated training dataset which is a time consuming and laborious task. 
The improved efficiency of UVid-Net by incorporating temporal information, along with reduced dependency on the availability of training data, will provide better segmentation of aerial videos. The lightweight architecture of UVid-Net aids in reducing the computational complexity and number of trainable parameters which makes it an ideal CNN architecture for UAV-based IoT applications. This improved segmentation can be utilized for monitoring of environmental changes, urban planning, disaster management and other aerial surveillance tasks.  In future, the developed system will be studied for real-time performance and be deployed in UAV drones for real-time scene analysis. 

In general, commercially available UAV are not flown with very high speed for applications such as scene analysis, surveillance etc.  The proposed model assumes a slow camera motion. In the presence of very large camera motion, there exists large scene variations between two consecutive frames. In these situations, estimation of temporal correspondence becomes mandatory for propagation of temporal information from frame to frame. However, the proposed work utilizing shot boundary detection and multi-branch encoder has shown to be robust to small camera motion. Moreover, the proposed approach is a more suitable method for UAV based IoT applications because of the reduction in the number of trainable parameters, computational complexity and transferable features.

{\small
	\bibliographystyle{ieee_fullname}
	\bibliography{egbib}
}
%

\vfill



\end{document}